\documentclass[preprint]{elsarticle}
\journal{Ocean Engineering}
\RequirePackage{amsmath}
\usepackage[margin=1in]{geometry}
\usepackage{lineno}
\usepackage{url}
\usepackage{comment}
\usepackage{subfig}
\usepackage{bbm}
\usepackage{bm}
\usepackage{amsmath,amssymb,amsfonts}
\usepackage{tensor}
\usepackage{esint}
\usepackage{algpseudocode}
\usepackage{graphicx}
\usepackage{textcomp}
\usepackage{adjustbox}
\usepackage{graphicx}
\usepackage{mdframed}\usepackage{xcolor}
\usepackage{mathtools,lipsum, nccmath,cuted}
\usepackage{siunitx}
\usepackage{textcomp}

\usepackage{comment}
\usepackage{makecell}
\usepackage{algorithm}
\usepackage{gensymb}
\usepackage{ulem}
\usepackage{colortbl}
\usepackage{enumitem}
\definecolor{Gray}{gray}{0.925}
\usepackage[textsize=tiny]{todonotes}

\graphicspath{ {figures/} }
\def\BibTeX{{\rm B\kern-.05em{\sc i\kern-.025em b}\kern-.08em
    T\kern-.1667em\lower.7ex\hbox{E}\kern-.125emX}}

\usepackage{pgf}
\usepackage{pgfplots}

\usetikzlibrary{shapes,arrows}
\usetikzlibrary{arrows.meta}
\usetikzlibrary{decorations.pathreplacing}
\usetikzlibrary{shapes.geometric}
\usetikzlibrary{shadows}

\usepackage{pgfplotstable}
\usepackage{graphicx}
\usepackage{multirow}
\usepackage{booktabs}
\pgfplotsset{compat=1.14}

\definecolor{blueLine}{RGB}{57,106,177}
\definecolor{blueFill}{RGB}{114,147,203}
\definecolor{redLine}{RGB}{204,37,41}
\definecolor{greenline}{RGB}{0,250,0}
\definecolor{blackLine}{RGB}{0,0,0}
\definecolor{goldLine}{RGB}{160,82,45}
\definecolor{goodGreen}{RGB}{213,232,212}
\definecolor{goodGreenBorder}{RGB}{150,192,129}
\definecolor{goodBlue}{RGB}{218,232,252}
\definecolor{goodBlueBorder}{RGB}{144,170,207}
\definecolor{goodPink}{RGB}{248,206,204}
\definecolor{goodPinkBorder}{RGB}{200,114,111}
\definecolor{airforceblue}{rgb}{0.36, 0.54, 0.66}
\definecolor{aquamarine}{rgb}{135, 206, 255}
\definecolor{deepskyblue}{rgb}{0.0, 0.75, 1.0}
\definecolor{persianblue}{rgb}{0.11, 0.22, 0.73}
\definecolor{aliceblue}{rgb}{0.94, 0.97, 1.0}

\usepackage[font=footnotesize]{caption}
\usepackage{subfig}
\usepackage{hyperref}

\makeatletter
\begin{document}
\begin{frontmatter}
\title{Sample-Efficient and Surrogate-Based\\Design Optimization of Underwater Vehicle Hulls}

\author[add1,add2]{Harsh Vardhan\texorpdfstring{\corref{corrauth}}{}}
\cortext[corrauth]{Corresponding author}
\ead{harsh.vardhan@vanderbilt.edu}
\author[add1,add2]{David Hyde}
\author[add2]{Umesh Timalsina}
\author[add2]{Peter Volgyesi}
\author[add1,add2,add3]{Janos Sztipanovits}
\address[add1]{Department of Computer Science, Vanderbilt University, 1400 18\textsuperscript{th} Ave.\ S., Nashville, TN 37212-2846, USA}
\address[add2]{Institute for Software Integrated Systems, Vanderbilt University, 1025 16\textsuperscript{th} Ave.\ S., Nashville, TN 37212-2328, USA}
\address[add3]{Department of Electrical and Computer Engineering, Vanderbilt University, 400 24\textsuperscript{th} Ave.\ S., Nashville, TN 37212, USA}
\date{}

\begin{abstract}
Physics simulations like computational fluid dynamics (CFD) are a computational bottleneck in computer-aided design (CAD) optimization processes.
To overcome this bottleneck, one requires either an optimization framework that is highly sample-efficient, or a fast data-driven proxy (surrogate model) for long-running simulations. Both approaches have benefits and limitations. 
Bayesian optimization is often used for sample efficiency, but it solves one specific problem and struggles with transferability; alternatively, surrogate models can offer fast and often more generalizable solutions for CFD problems, but gathering data for and training such models can be computationally demanding.
In this work, we leverage recent advances in optimization and artificial intelligence (AI) to explore both of these potential approaches, in the context of designing an optimal unmanned underwater vehicle (UUV) hull. 
For first approach, we investigate and compare the sample efficiency and convergence behavior of different optimization techniques with a standard CFD solver in the optimization loop.  For the second approach, we develop a deep neural network (DNN) based surrogate model to approximate drag forces that would otherwise be computed via the CFD solver.
The surrogate model is in turn used in the optimization loop of the hull design.
Our study finds that the Bayesian Optimization---Lower Condition Bound (BO-LCB) algorithm is the most sample-efficient optimization framework and has the best convergence behavior of those considered.
Subsequently, we show that our DNN-based surrogate model predicts drag force on test data in tight agreement with CFD simulations, with a mean absolute percentage error (MAPE) of 1.85\%.
Combining these results, we demonstrate a two-orders-of-magnitude speedup (with comparable accuracy) for the design optimization process when the surrogate model is used.
To our knowledge, this is the first study applying Bayesian optimization and DNN-based surrogate modeling to the problem of UUV design optimization, and we share our developments as open-source software.
\end{abstract}

\begin{keyword}
Bayesian optimization, surrogate modeling, design optimization, unmanned underwater vehicle, computational fluid dynamics.
\end{keyword}

\end{frontmatter}

\section{Introduction}
\label{sec:introduction}

Design optimization, including shape (geometry) and topology optimization, is a popular task in computational physics and engineering \cite{WANG2007395,HAZRA200546,VIQUERAT2021110080,ALEXANDROV2005121,AMSTUTZ2006573,ALLAIRE2004363}.
Although the promise of leveraging high-fidelity simulations is appealing for CAD, an obvious roadblock is the curse of dimensionality associated with simulating complex PDE-governed multiphysics systems (e.g., the $O(n^4)$ (space plus time) scaling of three-dimensional Eulerian fluid simulations).
While an obvious workaround is to use cheaper and less accurate simulation models, a more prudent strategy for design optimization is to devise a highly sample-efficient optimization framework (such that few simulations need to be run to find an optimal design) or to construct a surrogate model that can efficiently approximate the behavior of highly accurate simulations.
In particular, surrogate modeling is made far more powerful due to recent advances in AI and deep learning.

We pursue both sample-efficienct optimization and surrogate modeling approaches for a real-world problem of optimizing the design of a UUV.
UUVs have a broad scope of both civilian and military applications, ranging from oceanography to payload delivery, anti-submarine warfare, and reconnaissance, to name a few~\cite{allard2014unmanned,fletcher2000uuv,button2009survey}.
Such applications require long-range endurance and the operation of the UUV.
For more energy-efficient and longer-range UUVs, it is critical to optimize the shape of the hull, which has the most dominant effect on the drag resistance arising from its movement through the water.
Hull designs focusing on an axisymmetric body of revolution are the most popular due to their hydrodynamic and homeostatic stability, pressure-bearing capacity, and space utilization.
Axisymmetric bodies of revolution used in UUVs hulls are classified into various types~\cite{liu2021fine}, of which the Myring-type hull profile \cite{myring1976theoretical} is most commonly used due to its optimized inner space, minimized drag force, streamlined flow characteristics, and favorable geometry for both dynamic and hydrostatic pressure.
Many commercial UUVs today like Blufin, Remus100, and Swordfish use a Myring hull design \cite{french2010analysis}. 
Due to their popularity \cite{french2010analysis}, we focus on Myring hull designs in this work, though our methodology is general.

In the first part of the present work, our goal is to investigate different optimization and sampling methods' performance with a CFD simulator in the loop to find the optimal shape of a UUV hull.
We compare several numerical methods, selected based on the authors' knowledge of current practice.
For comparing the performance of these different methods, we are interested in two properties: \textit{expected sample efficiency} in finding the optimal design, and \textit{variance in the search process} (convergence behavior).
These are defined as follows:
let $S$ be the set of labeled sample designs in the process of design optimization, and let $f$ map each candidate hull design $S_i$ to the steady-state drag force $F$ experienced as the UUV with design $S_i$ flows through water (defined more carefully in Section \ref{sec:Methodology}).
A sample-efficient design optimization process minimizes the expected cardinality of $S$ while also discovering an optimal design (i.e., finding an argmin of $f$, whose minimum is $F^\text{optimal}$).
Although $F^\text{optimal}$ is not known a priori, we can find an observed optimum by minimizing all evaluations obtained by running different optimization methods.
Since each optimization process is different in such cases, we are interested in expected sample efficiency, i.e., the method that yields the most optimal design in expectation on a given budget.
Variability in the search process (convergence behavior) measures the variance in the optimization process when started with different initializations in design space.
We measure variance by $|F^\text{max}- F^\text{min}|$, i.e., the range of drag coefficients observed across each iteration.
To our knowledge, there is no empirical study on the convergence performance and sample efficiency of the methods we study in the context of CFD simulations.
We ultimately find that Bayesian Optimization---Lower Confidence Bound (BO-LCB) is the best-suited method for our considered UUV design optimization problem.

In the second part of this work, we train an AI-based surrogate on a high-fidelity dataset generated by Reynolds-averaged Navier-Stokes (RANS) simulation of various UUV designs moving through an incompressible flow field.
For data generation, we evaluated 3,021 different designs sampled from the design space of a small UUV class \cite{crowell2013design}.
The motivation for training an AI surrogate is to leverage the generalization power of the trained surrogate model to replace the computationally expensive numerical simulation process without a significant loss in accuracy.
Specifically, we seek drag force values inferred from the trained AI surrogate that are in good agreement with those obtained via a numerical solver (yet which are obtained at a very low cost).
We succeed on this front, obtaining a MAPE of $1.85\%$.
We later use this trained surrogate for surrogate-based optimization and observe that the optimal design obtained on a given budget is similar in quality to running CFD in the loop.
However, surrogate-based optimization takes only a few milliseconds to find an optimum, instead of the standard CFD-in-the-loop procedure that takes hours to find this design.

Throughout the paper, for CFD simulation, we use the Reynolds-averaged Navier-Stokes (RANS) flow model \cite{sagaut2013multiscale}, along with a $k$-$\omega$ shear stress transport (SST) \cite{menter1992improved} turbulence model.
We use the implementations provided in OpenFOAM \cite{jasak2007openfoam}.
We have freely released the source code for the paper, including for running CAD-CFD simulations, running optimization on these designs, and our surrogate model with data and trained weights, at \url{https://github.com/vardhah/UUV-design-optimization}.

The contributions of this paper are summarized as follows:
\begin{itemize}
	\item We provide a comprehensive evaluation of a number of optimization methods on a real-world design optimization problem, which is useful for practitioners considering adopting these algorithms.
	\item We demonstrate that a neural network can accurately capture a complex property of RANS-based fluid dynamics (drag force/drag coefficient).
	\item We show that a sample-efficient, surrogate-enhanced design optimization algorithm can find optimal UUV hull designs in milliseconds, representing a two-orders-of-magnitude speedup over design optimization loops that use RANS numerical simulation.  Moreover, we are not aware of any previous studies that leverage an AI model for drag in the context of RANS with $k$-$\omega$ SST to optimize UUVs.
	\item We provide a ready-to-use software package for drag-based design optimization that can be readily generalized to other physical scenarios.
\end{itemize}

The remainder of the manuscript is structured as follows.
After a discussion of related work in Section \ref{sec:related_work},
Sections \ref{sec:background} and \ref{sec:Methodology} present our methodology, CAD design, numerical setting, and mathematical models used for CFD simulation.
In Section \ref{sec:DO}, we study the different direct optimization methods deployed for a UUV design use case and explain our findings on the performance of all used optimization methods.
In Section \ref{sec:ga_sbo}, we set up the problem of surrogate modeling as our goal with the aim to solve a large class of problems.
We explain our data generation process, our learning architecture, training details, and the performance of the trained model on test data.
We also show the results and speedups of our surrogate model when used in an optimization loop (surrogate-based optimization) in comparison to CFD-in-the-loop optimization.
Concluding remarks are provided in Section \ref{sec:conclusion}.
 \section{Related Work}
\label{sec:related_work}

Optimizing UUV hull design for low drag is a task that was considered as early as 1950, originally being studied experimentally.
Gertler \cite{gertler1950resistance} designed 24 different bodies (known as ``Series 58'') using five parameters and a polynomial equation that define the shape of the UUV body. These 24 different hull designs were towed through the water at different speeds to measure drag. 
Later, in another experimental study, Carmichael \cite{carmichael1966underwater} showed that by reducing the fineness ratio or surface area, drag is reduced for a constant frontal area or constant volume applications.
Carmichael designed a tail-boomed body (``Dolphin'') and tested it extensively in the Pacific Ocean, showing a 60\% drag reduction compared to conventional torpedo shapes of equal volume.
More recent reviews on UUV design can be found in Neira et al.\ \cite{neira2021review} and Manley \cite{manley2016unmanned}.

Theoretical advances and development of numerical methods for simulating UUVs contributed to significant progress in UUV design over the intervening decades.
Parsons et al.\ \cite{parsons1974shaping} were among the first to develop a computer-oriented optimization procedure for an axisymmetric body of revolution in a finite constrained parameter space, but among other limitations, the method only applied to laminar flow.
Hertel \cite{hertel1966full} investigated fuselage shape with maximum volume and considered fast-swimming animals.
His investigation showed that bodies of revolution of profile identical to NACA two-dimensional laminar shapes do not provide the proper pressure gradient for the boundary layer to remain laminar, and in contrast, parabolic-nose bodies like Dolphin and Shark provide a more streamlined boundary layer.
In 1976, Myring \cite{myring1976theoretical} introduced boundary layer calculations based on the viscous-inviscid flow interaction method to predict drag on the body of revolution at Reynolds number $10^7$.
His study concluded that body drag is less variable when the nose or tail varies from slender to stout within a certain range, but that drag increases dramatically once outside that range; this yielded the so-called Myring hull class of designs.
At the same time, Hess \cite{hess1976problem} developed a simplified integral drag formula and used it to compare drag performance over a wide variety of bodies.
Zedan and Dalton \cite{zedan1979viscious} studied drag characteristics of a number of axisymmetric body shapes, concluding that an unconventional laminar-shaped body is the best candidate at a high Reynolds number for the lowest drag design.
Lutz et al.\ \cite{lutz1998drag} developed a numerical shape optimization method and implemented a linear stability theory to estimate the drag of axisymmetric bodies.
Using this they designed a minimal drag design for a given inner space. A first-order Rankine panel method is developed by Alvarez et al.\ \cite{alvarez2009hull} to optimize the hull shape of a UUV operating under snorkeling conditions near a free surface.
More recent works applying CFD analysis to UUV design include Stevenson et al.\ \cite{stevenson2007auv}, Wei et al.\ \cite{zifan2014analysis}, and Yamamoto \cite{yamamoto2015research}, which together suggest that various UUV designs can be optimal under different flow conditions---there is not a universal optimal UUV design.

With the advent of computer-aided design, traditional CFD can be leveraged inside an optimization loop in order to seek an optimal UUV design for given flow conditions, see e.g.\ Alam et al.\ \cite{alam2015design}.
Many different optimization algorithms have been considered.
For instance, adjoint methods \cite{jameson2003aerodynamic} and genetic algorithms \cite{song2010research} can be applied.
Schweyher et al.\ \cite{schweyher1996optimization} used an evolutionary strategy to obtain a minimum-drag body.
The application of Bayesian optimization to finding a minimum-drag shape is studied in \citet{vardhan2023search, eismann2017shape}, and  \citet{vardhan2023constrained}. However, these studies did not conduct any comparative study on the performance of different optimization methods. Furthermore, some of these studies used Expected Improvement (EI) \cite{eismann2017shape} as an acquisition function, which we find (reported in our results section) has disappointing convergence properties. Vardhan et al.\ \cite{vardhan2022data} studied a similar problem setting but with the goal of not optimizing but testing their active learning algorithm in the context of engineering problems. 
Despite these studies, underwhelming adoption of CFD-in-the-loop design optimization techniques continues, due in large part to the computational expense of running high-fidelity CFD simulations at each iteration of the optimization loop.

To address this limitation, over the last few decades, surrogate modeling and surrogate-based optimization has been used as a substitute for complex engineering simulation processes, using e.g.\ Kriging/Gaussian Process \cite{rasmussen2003gaussian} based learning models or other polynomial equations based data fit models \cite{forrester2008engineering,marsden2004optimal,abouhussein2023computational}. Gaussian Process Regression (GPR) methods offer sample efficiency due to their complex internal model based approach for single-instance problem solving, often yielding good results within approximately 100 evaluations \cite{booker1999rigorous,morita2022applying}. However, a significant limitation lies in the non-transferability of these evaluated samples to other similar optimization problems.  In contrast, one may develop a surrogate model that covers a wider range of the design space and exhibits stronger generalization capabilities.  Such a model enables addressing multiple design challenges and acts as a digital twin of the design evaluation process, which one can evaluate more rapidly compared to evaluating each  problem via CFD.
Recent research has focused on developing surrogate models for accurately approximating flow field behavior. This includes investigating how fluid flow impacts the shape and aerodynamics of objects immersed in the fluid.
Bhatnagar et al.\ \cite{bhatnagar2019prediction} designed and trained a convolutional-based neural network architecture in an encode-decoder fashion to predict the flow characteristics of aerodynamics flow fields on 2D airfoil shapes.
Chen et al.\ \cite{chen2019u} used a U-net \cite{ronneberger2015u} based learning architecture to predict 2D velocity and pressure fields around arbitrary shapes in laminar flows.
Machine learning-based surrogate models are also designed in various other engineering domains to speed up design discovery \cite{vardhan2021machine,vardhan2022deep}. 
In all these studies, a learning model is trained to either learn the flow field directly or to construct a mapping between shape parameters and the quantity of interest (e.g., drag or lift force).

There are multiple aspects where these studies leave opportunities for improvement.
First, the physics used for flow evolution is either a laminar or simple turbulence model like $k$-$\omega$ or $k$-$\epsilon$.
Real-world UUV design problems need more complex turbulence modeling; for instance, one study on the design of real-world UUVs \cite{jones2016rans} found that the $k$-$\omega$ SST turbulence model predicted closest to the ground-truth data from their experiments.
We did not find works that attempted to approximate flow behavior or its effect while leveraging $k$-$\omega$ SST as a turbulence model.
Second, the aforementioned works on CFD surrogate modeling either used small airfoils and 2D shapes, which do not reflect the actual complexity involved in surrogate creation for real-world designs.  In the realm of 3D real-world design scenarios, CFD surrogate modeling encounters several significant challenges. These include complexities in mesh generation, which affect the accuracy and resolution of the model. Additionally, the computational burden is notable, particularly in terms of the time required to evaluate each sample. There is also a practical limit on the number of samples that can be generated within a reasonable timeframe, which can impact the comprehensiveness and reliability of the surrogate model. \section{Background}
\label{sec:background}

\subsection{Governing Equations and Turbulence Modeling}

The objective of the first part of the present work is to use CFD simulations in an optimization loop to discover UUV hull shapes with minimal drag.
In order to analyze turbulent fluid flow and drag on a given candidate design, we first choose a model for the fluid.
The Navier–Stokes equations for Newtonian, incompressible, isothermal fluid can be described as
\begin{gather}
\rho \frac{d \vec{v}}{dt} = -\nabla p + \mu \nabla^2 v + \rho \vec{g} \\
\nabla \cdot \vec{v} = 0 .
\end{gather}
Here, $\vec{v}$ is the velocity field, $\mu$ is the viscosity of the fluid, $p$ is the pressure, and $\vec g$ is the gravitational acceleration vector.
We assume a constant fluid density $\rho$, which is generally a suitable assumption for water along with incompressibility. 
To efficiently approximate direct numerical simulation of the Navier-Stokes equations, common choices are Reynolds-averaged Navier–Stokes (RANS) with a closure-based turbulence model and large eddy simulation (LES) \cite{sagaut2013multiscale}.
In this work, we use a RANS model that solves time-averaged mass and momentum conservation equations \cite{jones2016rans}.

For turbulence physics, we use the $k$-$\omega$ shear stress transport (SST) model \cite{menter1992improved}.
This SST-based closure model relies on the Boussinesq hypothesis on the Reynolds stress tensor and includes a modified definition of the scalar eddy viscosity coefficient.
The underlying assumption is that the Reynolds stress tensor is related to the mean rate-of-strain (deformation) tensor and the turbulent kinetic energy $k$.
It uses two different turbulence models in different regions of the flow: the $k$-$\omega$ model \cite{wilcox1998turbulence} is used in the near wall region, and the $k$-$\epsilon$ model \cite{launder1983numerical} is used in the free stream region that is away from the boundary layer.
Further, this model also takes into account the transport of the principal shear stress in adverse pressure gradient boundary layers.
The model mitigates sensitivity to inlet free-stream turbulence parameter values, which is a limitation of the $k$-$\omega$ model \cite{menter1992influence}.
Due to all these benefits, the $k$-$\omega$ SST closure model has had great success and has been validated against various real-world turbulent flow problems \cite{menter1992improved,tide2008comparison,nikiforow2016designing}.
The mathematical model of the $k$-$\omega$ SST model used in OpenFOAM is given by \cite{menter1992improved}.

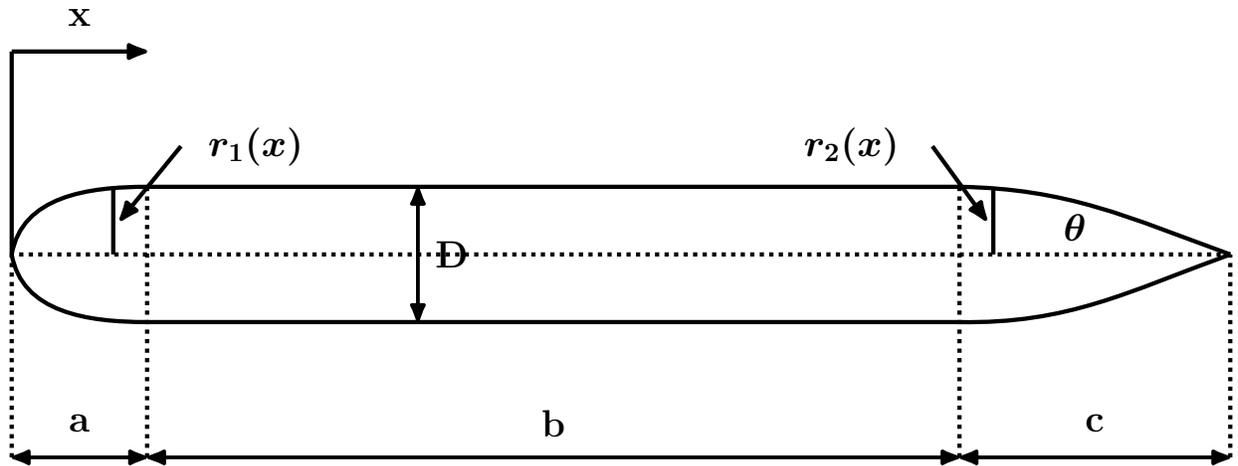
\begin{figure}[ht!]
    \centering
\begin{tikzpicture}[scale=0.9,]
\centering

\tikzstyle{arrow} = [ultra thick, line width=0.5mm,{{Latex[round]}-{Latex[round]}}, align=center]

\coordinate (a) at(7,-30) {};

\coordinate (b) at(3,-30) {};
\coordinate (c) at(-9,-30) {};

\coordinate (e) at(3,-28) {};
\coordinate (f) at(-9,-28) {};

\coordinate (g) at(3,-26) {};
\coordinate (h) at(-9,-26) {};

\coordinate (i) at(7,-27) {};
\coordinate (j) at(-9,-27) {};

\coordinate (k) at(-9,-27) {};
\coordinate (l) at(-11,-27) {};
\coordinate (m) at(-11,-24) {};
\coordinate (n) at(-9,-24) {};
\coordinate (d) at(-11,-30) {};

\draw [arrow] (a) -- (b);
\draw [arrow] (b) -- (c);
\draw [arrow] (c) -- (d);

\draw [arrow] (-5,-28) -- (-5,-26);
\node[draw=none, ultra thick]at (-4.5, -27) {{\textbf{\Large{D}}}};

\draw [ultra thick] (e) -- (f) ;
\draw [ultra thick] (g) -- (h);
\draw [ultra thick] (-9.5,-27) -- (-9.5,-26);
\draw [ultra thick] (3.5,-27) -- (3.5,-26);

\node[draw=none, ultra thick]at (-7.5, -25.4) {{\textbf{\Large{ $\bm{r_1(x)}$}}}};
\draw [ultra thick, -{Latex[round]}](-8.5, -25.4) -- (-9.4, -26.5);

\node[draw=none, ultra thick]at (1.3, -25.4) {{\textbf{\Large{ $\bm{r_2(x)}$}}}};
\node[draw=none, ultra thick]at (4.6, -26.6) {{\textbf{\Large{ $\bm{\theta}$}}}};
\draw [ultra thick, -{Latex[round]}](2.6, -25.4) -- (3.4, -26.5);

\draw [dotted, ultra thick](i) -- (j);

\draw [dotted, ultra thick](c) -- (h);
\draw [dotted, ultra thick](b) -- (g);
\draw [dotted, ultra thick](a) -- (i);
\draw [dotted, ultra thick](k) -- (l);
\draw [dotted, ultra thick](d) -- (l);
\draw [ultra thick](l) -- (m);
\draw [ultra thick, -{Latex[round]}](m) -- (n);

\draw [ultra thick](e) to[out=-1,in=200](i);
\draw [ultra thick](g) to[out=-1,in=-200](i);

\draw [rounded corners=10mm,ultra thick](f) to[out=-180,in=-80] (l);
\draw [rounded corners=10mm,ultra thick](h) to[out=-180,in=80] (l);
\node[draw=none, ultra thick] at (-10, -29.5){{\textbf{\Large{a}}}};
\node[draw=none, ultra thick] at (-3,-29.5) {\textbf{\Large{b}}};
\node[draw=none, ultra thick] at (5,-29.5) {\textbf{\Large{c}}};
\node[draw=none, ultra thick] at (-10,-23.5) {\textbf{\Large{x}}};

\end{tikzpicture}     \caption{A typcial Myring Hull profile.  The bow/nose (left) and stern/tail (right) have independently parameterized profiles, and the overall design is axisymmetric as indicated by the horizontal dashed line.}
    \label{fig:myring}
\end{figure}

\subsection{Myring Hulls}

\begin{figure}[ht!]
        \centering
\begin{tikzpicture}[scale=0.85,]
\centering

\tikzstyle{arrow} = [ultra thick, line width=0.5mm,{{Latex[round, length=5mm, width=3mm]}-{Latex[round,length=5mm, width=3mm]}}, align=center]

\tikzstyle{arrow1} = [ultra thick, line width=0.5mm,{-{Latex[round,length=5mm, width=2mm]}}, align=center]

\coordinate (a) at(7,-30) {};

\coordinate (b) at(-1,-30) {};
\coordinate (c) at(-5,-30) {};
\coordinate (d) at(-12,-30) {};
\coordinate (e) at(7,-29) {};
\coordinate (f) at(-12,-29) {};

\coordinate (g) at(-0.5,-26.5) {};
\coordinate (h) at(-5.5,-26.5) {};

\coordinate (i) at(-1,-26.5) {};
\coordinate (j) at(-1,-30.5) {};

\coordinate (k) at(-5,-26.5) {};
\coordinate (l) at(-5,-30.5) {};
\coordinate (m) at(-12,-29.3) {};
\coordinate (n) at(-12,-30.5) {};
\coordinate (o) at(7,-29.3) {};
\coordinate (p) at(7,-30.5) {};
\coordinate (q) at(-3,-26.5) {};
\coordinate (r) at(-3,-29) {};

\draw [arrow] (a) -- (b);
\draw [arrow] (b) -- (c);
\draw [arrow] (c) -- (d);
\draw [ultra thick] (e) -- (f) ;
\draw [ultra thick] (g) -- (h);
\draw [ultra thick](i) -- (j);
\draw [ultra thick](k) -- (l);
\draw [ultra thick](m) -- (n);
\draw [ultra thick](o) -- (p);
\draw [arrow] (q) -- (r);
\node[draw=none, ultra thick] at (-8.5, -29.5){{\textbf{\Large{a}}}};
\node[draw=none, ultra thick] at (-3,-29.5) {\textbf{\Large{b}}};
\node[draw=none, ultra thick] at (3,-29.5) {\textbf{\Large{c}}};
\node[draw=none, ultra thick, rotate=90] at (-3.7,-27.8) {\textbf{\Large{ $\bm{D/2}$ }}};

\draw [ultra thick] (-0.7,-26.5) to[out=-15,in=-180] (e);
\draw [ultra thick](-0.7,-26.5) to[out=-5,in=-180] (e);
\draw [ultra thick](g) to[out=5,in=-180] (e);

\draw [arrow1] (-.8,-28.5) -- (1, -28.5)-- (1.6, -27.5);
\node[draw=none, ultra thick] at (.1,-28.2) {\small{{ $\bm{\theta=5^{\circ}}$ }}};

\draw [arrow1] (3.8,-26) -- (2, -26)-- (1.4, -27);
\node[draw=none, ultra thick] at (3,-25.7) {\small{{ $\bm{\theta=15^{\circ}}$ }}};

\draw [arrow1] (5.8,-26.7) -- (4, -26.7)-- (3.4, -27.7);
\node[draw=none, ultra thick] at (5,-26.4) {\small{{ $\bm{\theta=25^{\circ}}$ }}};
\draw [ultra thick](f) to[out=17,in=180] (h);
\draw [ultra thick](f) to[out=7,in=180] (h);
\draw [ultra thick](-10.2,-29) to[out=7,in=180] (h);

\draw [arrow1] (-5.2,-28.5) -- (-7, -28.5)-- (-7.6, -27.5);
\node[draw=none, ultra thick] at (-6.1,-28.2) {\small{{ $\bm{n=0.5}$ }}};

\draw [arrow1] (-11.2,-26.5) -- (-9.6, -26.5)-- (-9, -27.5);
\node[draw=none, ultra thick] at (-10.4,-26.2) {\small{{ $\bm{n=2.0}$ }}};

\draw [arrow1] (-12.2,-27.3) -- (-10.6, -27.3)-- (-10, -28.3);
\node[draw=none, ultra thick] at (-11.5,-27) {\small{{ $\bm{n=1.5}$ }}};

\end{tikzpicture}         \caption{Demonstrating the influence of $n$ and $\theta$ on the shape of the upper portion of a Myring hull profile.  Larger $n$ and $\theta$ tend to increase the volume of the nose and tail, respectively.}
        \label{fig:ntheta}
\end{figure}
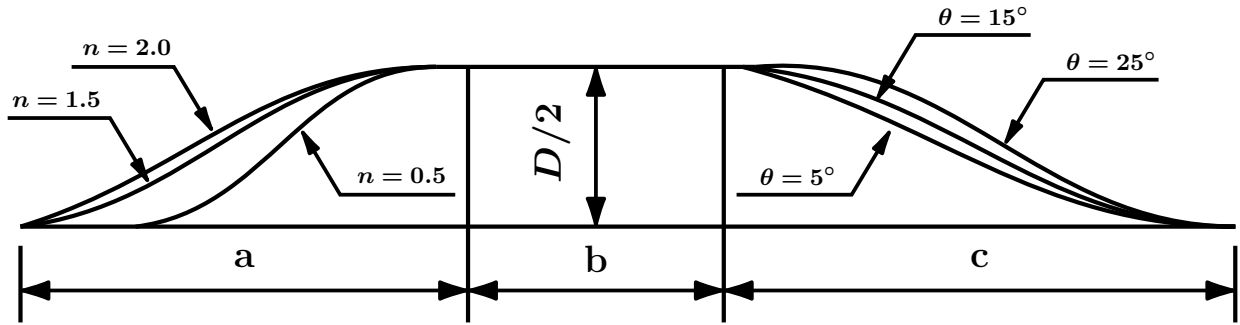

From a design perspective,  hull geometry has a dominant effect on the drag of the vehicle. Out of many modern hull designs, the Myring-type hull profile \cite{myring1976theoretical} is most commonly used.
It has an axisymmetric body of rotation (a cylindrical body with a large ratio between the length to diameter) and has a number of advantages such as optimized inner space, minimized drag force, streamlined flow characteristics, and favorable geometry for both dynamic and hydrostatic pressure.
We focus on optimizing Myring hull designs in the present work.
The bow and stern equations for Myring hull profiles (refer to Figure \ref{fig:myring}) are given by:
\begin{gather}
    r_1(x) = \frac{1}{2}D\Big[1-\Big(\frac{x-a}{a}\Big)^2\Big]^{\frac{1}{n}} \\[3ex]
    r_2(x) = \frac{1}{2}D -\Big[ \frac{3D}{2c^2}-\frac{\tan \theta}{c}\Big](x-a-b)^2+\Big[ \frac{D}{c^3}-\frac{\tan \theta}{c^2}\Big](x-a-b)^3 .
\end{gather}
Here, $r_1$ and $r_2$ are parameterized functions that define the profile of the nose/bow and the tail/stern of the hull, respectively.
$x$ is measured from the tip of the nose.
$D$ is the diameter of the hull, and $a$, $b$, and $c$ are the nose length, body length, and tail length, respectively (refer to Figure \ref{fig:myring}).
$n$ and $\theta$ are parameters that define the shape and tapering of the nose and tail in different streamlined profiles.
By changing the values of $n$ and $\theta$, the profile of the body of the hull can be controlled.
Figure \ref{fig:ntheta} shows the effect of changing $n$ and $\theta$ on the shape of the hull. The effect of nose and tail offset is ignored in this study. A variety of shapes can be generated by the combination of these six parameters, $a$, $b$, $c$, $D$, $n$, and $\theta$.
As seen in the next section, we can perform CFD-in-the-loop design optimization by parameterizing a 3D mesh on these values to generate candidate designs.

\section{Design Loop Overview and Simulation Setup}
\label{sec:Methodology}
\begin{figure}[ht!]
    \centering
    \begin{tikzpicture}[scale=0.9]
\node (box) [draw=none, dashed, ultra thick, minimum height = 9cm, minimum width = 17cm] at (0, 0){};

\node (docker1container) [below of=box, xshift=-5cm, yshift=2.9cm,fill=blue!10 ,draw=black, dashed, thin, minimum height = 5.15cm, minimum width=7.0cm] at (0, 0){};

\node (docker1text) [below of=docker1container, xshift=0cm, yshift=3.3cm, align=center]{\textbf{\textit{CAD Docker Container}}};

\node (docker2container) [below of=box, xshift=5cm, yshift=2.9cm, fill=green!10,draw=black, dashed, thin, minimum height = 5.15cm, minimum width=7.0cm] at (0, 0){};

\node (docker1text) [below of=docker2container, xshift=0cm, yshift=3.3cm, align=center]{\textbf{\textit{CFD Docker Container}}};

\node (docker1container) [below of=box, xshift=0 cm, yshift=-1.9cm,fill=red!5 ,draw=black, dashed, thin, minimum height = 3.15cm, minimum width=17.0cm] at (0, 0){};

\node (docker1text) [below of=docker1container, xshift=-4.0cm, yshift=0cm, align=center]{\textbf{\textit{Python environment and scripts}}};

\node (cadseed) [below of=box, minimum height=2cm, minimum width=2cm, align=center, draw=black, ultra thick, xshift=-7cm, yshift=1.7cm] {CAD Seed \\ Design};

\node (meshing) [below of=box, dashed, minimum height=2cm, minimum width=2cm, align=center, draw=black, ultra thick, xshift=-3.5cm, yshift=3.7cm] {CAD Design + Body\\ Meshing + STL \\ process};

\node (cadtool) [below of=box, minimum height=2cm, minimum width=2cm, align=center, draw=black, ultra thick, xshift=-3.5cm, yshift=1.7cm] {CAD Design\\ Tool (FreeCAD)};

\node (stlfile) [below of=box, minimum height=1cm,, minimum width=1cm, align=center,fill={rgb:orange,1;yellow,2;pink,5}, draw=black, ultra thick, xshift=0cm, yshift=1.7cm] {STL File};

\node (rans) [below of=box, dashed, minimum height=2cm, minimum width=3.3cm, align=center, draw=black, ultra thick, xshift=3.5cm, yshift=3.7cm] {Meshing + RANS \\ + $k$-$\omega$ SST};

\node (cfd) [below of=box, minimum height=2cm, minimum width=2cm, align=center, draw=black, ultra thick, xshift=3.5cm, yshift=1.7cm] {CFD \\Simulation\\ (OpenFOAM)};

\node (dragforce) [below of=box, minimum height=1cm, minimum width=1cm, align=center, draw=black, ultra thick, xshift=7cm, yshift=1.7cm] {Drag Force};

\node (init) [below of=box, minimum height=2cm, minimum width=2cm, align=center, draw=black, ultra thick, xshift=3.5cm, yshift=-1.5cm] {Initialization \\and Boundary\\ Conditions};

\node (chosenparams) [below of=box, minimum height=2cm, minimum width=2cm, align=center, draw=none, ultra thick, xshift=-1.8cm, yshift=-0.1cm] {Chosen parameters};

\node (randomsampling) [below of=box, minimum height=2cm, minimum width=2.0cm, align=center, draw=black, ultra thick, xshift=-1.8cm, yshift=-1.5cm] {Sampler\\ and\\ Optimizer };

\node (datacollect) [below of=box, minimum height=2cm, minimum width=2.5cm, align=center, draw=black, ultra thick, xshift=0.5 cm, yshift=-1.5cm] {Data\\ Collection};

\node (bds) [below of=box, minimum height=2cm, minimum width=2cm, align=center, draw=black, ultra thick, xshift=-7cm, yshift=-1.5cm] {Bounded\\ Design \\Space };

\draw [ultra thick, -stealth](cadseed) -- (cadtool);
\draw [ultra thick, -stealth](cadtool) -- (stlfile);
\draw [ultra thick, -stealth](stlfile) -- (cfd);
\draw [ultra thick, -stealth](cfd) -- (dragforce);
\draw [ultra thick, -stealth](init) -- (cfd);
\draw [ultra thick, -stealth](bds) -- (randomsampling);
\draw [ultra thick, -stealth](dragforce.south) -- ++(0, -5cm) -- ++(-7.2cm, 0) -- (datacollect);
\draw [ultra thick, -stealth](datacollect.north) -- ++(0, 0.7cm) -- ++(-4.4cm, 0) -- (cadtool.south);

\end{tikzpicture}     \caption{An integrated toolchain incorporating FreeCAD (parametric CAD modeling software) and OpenFOAM (CFD simulation software) in a Python environment (controlling the process flow and running optimizers and samplers) for CAD design generation, drag evaluation, and optimization.} \label{fig:tool_chain}
\end{figure}
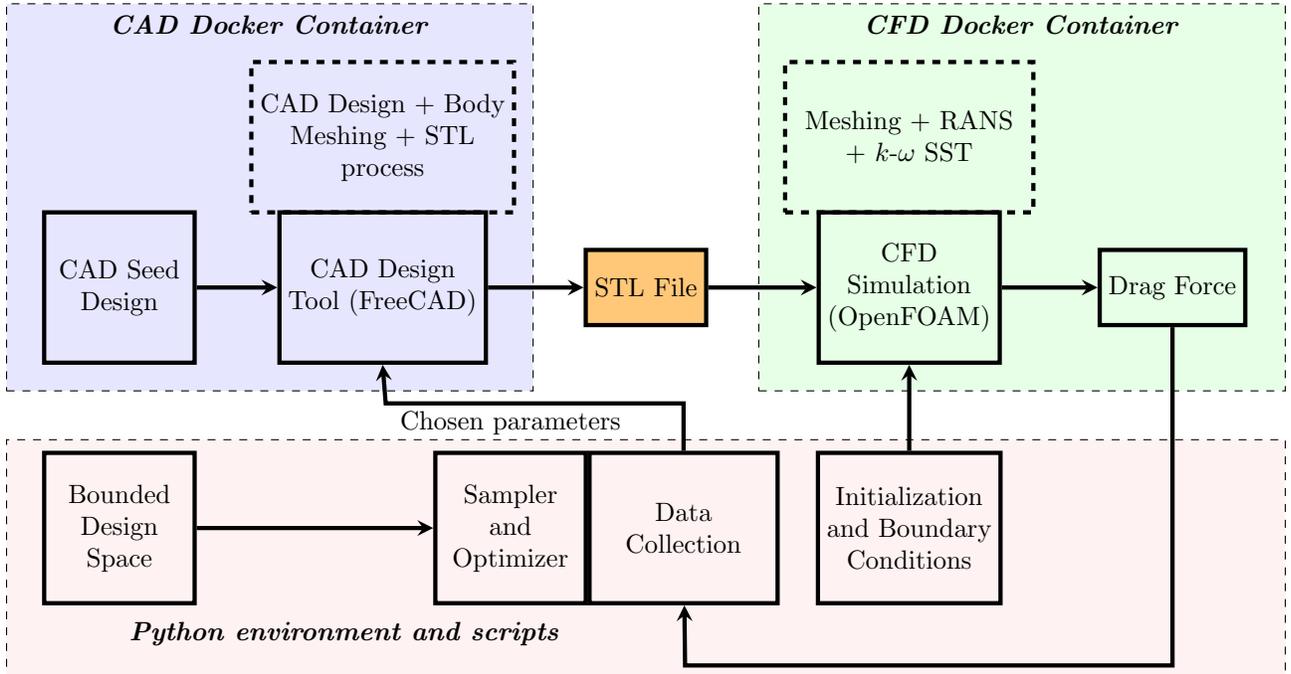

We build an end-to-end software pipeline for automated design space exploration and optimization. The UUV hull design process involves two engineering design/simulation tools and one optimization/sampling framework.
The design and simulation tools we use are FreeCAD \cite{riegel2016freecad}, a parametric CAD modeling tool, and OpenFOAM \cite{jasak2007openfoam}, a C++-based platform for fluid physics simulations via finite volume discretizations.
OpenFOAM also consists of auxiliary tools for meshing, parallelism, and linear solvers.
For optimization and sampling purposes, we choose Python-based implementations of various optimization and sampling algorithms due to easy deployment, extensibility, and availability of open-source software packages.
These three tools are combined as an integrated toolchain (refer to Figure~\ref{fig:tool_chain}).
For ease of versioning, maintenance, and reproducibility, we use Docker as a packaging and software delivery tool for both FreeCAD and OpenFOAM.
Details of the organization of our code are available in the online repository.

For parametric design optimization and data generation, we begin with a parametric 3D CAD seed design, whose range of parameters determines the spectrum of possible UUV hull models we consider (the design space).
For a given set of parameter values, our tooling invokes FreeCAD, which automatically designs a 3D surface geometry, triangulates the surface, and saves it as an STL file. 
The generated STL file is then transferred to another Docker container, which is prepared with OpenFOAM for running the fluid simulation on the given STL design.
In addition to the candidate design STL file, the OpenFOAM Docker container is provided various settings from our Python scripts like boundary conditions and solver parameters.
Once the OpenFOAM simulation converges, the Docker container returns the steady-state drag force measured on the input design to our Python scripts, which can then perform an update in an optimization algorithm.
For CFD-in-the-loop optimization, we used the GPyOpt \cite{gpyopt2016}, pyMOO \cite{blank2020pymoo}, and pyDOE software packages for optimization algorithms and sampling methods.
During the data collection process for surrogate modeling, we used uniformly random design points in the design space. We note that in general, ensuring valid model outputs from parametric CAD design is challenging \cite{hoffmann2001towards}, and we ensured model validity (avoiding inverted meshes, etc.) by manually inspecting generated models and limiting parameter values to within reasonable bounds (see Table \ref{tab:ds}).

\begin{figure}[ht!]
    \centering
    \label{fig:meshing}
    \subfloat[]{
        \includegraphics[width=0.49\textwidth]{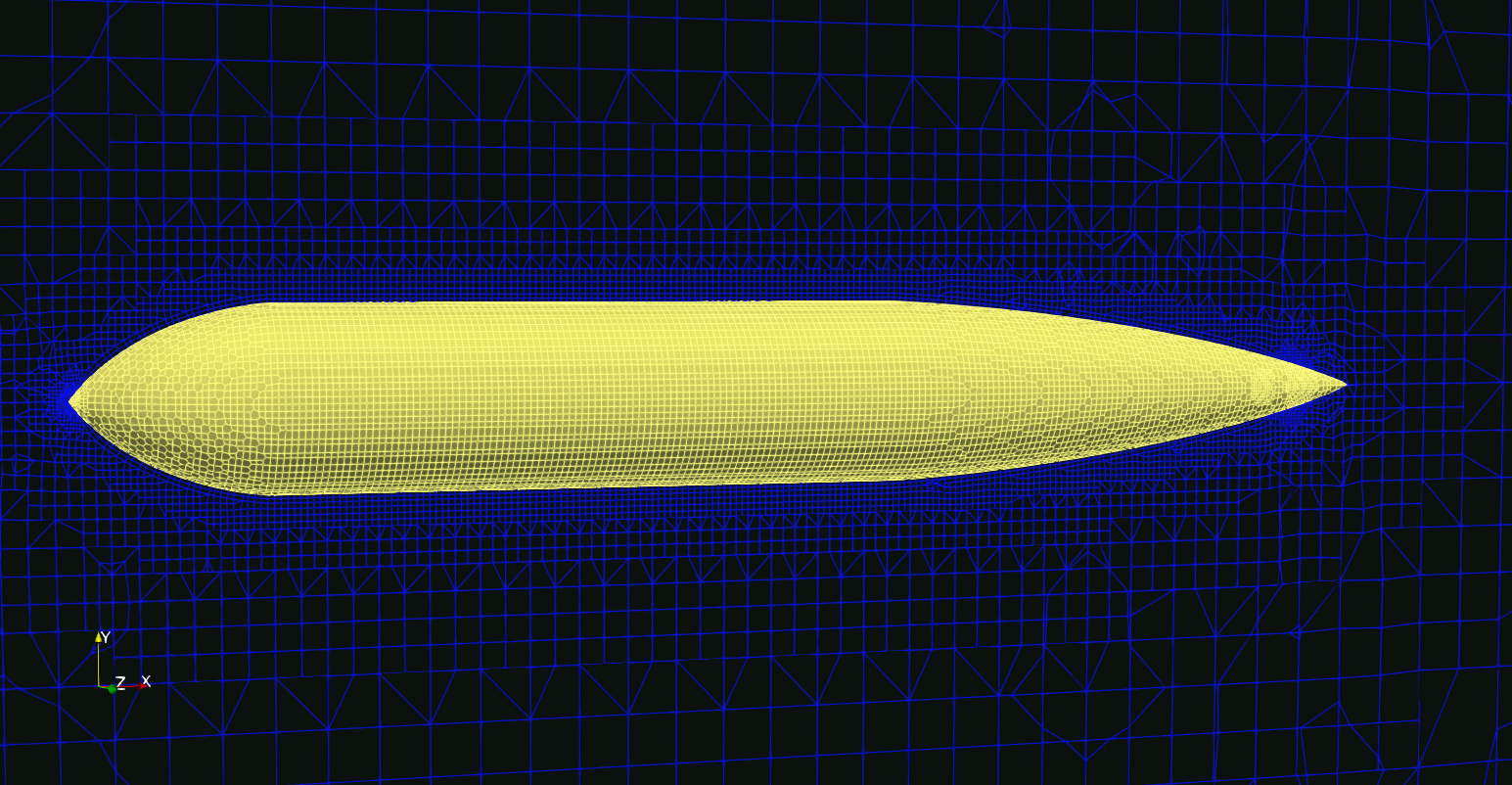}
        \label{fig:meshing:cross}
    }
    \subfloat[]{
        \includegraphics[width=0.49\textwidth]{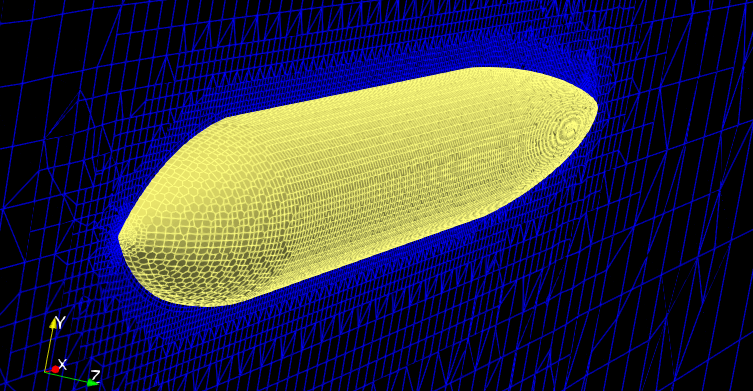}
        \label{fig:meshing:oblique}
    }
    \caption{Meshing in OpenFOAM after \texttt{blockMesh}-based volumetric mesh generation and \texttt{snappyHexMesh}-based refinement. \protect\subref{fig:meshing:cross} Cross-sectional view, \protect\subref{fig:meshing:oblique} Oblique view. OpenFOAM performs adaptive meshing with a maximum number of allowable cells provided as a parameter by the user (in our experiments, 1 million).}
\end{figure}

We note that OpenFOAM's solver expects a volumetric rather than a surface mesh.
We use OpenFOAM's built-in \texttt{blockMesh} and \texttt{snappyHexMesh} tools to create a volumetric mesh from the surface mesh generated by FreeCAD.
The volumetric mesh is composed of hexahedra and split-hexahedra elements.
Figures \ref{fig:meshing:cross} and \ref{fig:meshing:oblique} provide a view of the mesh in the vicinity of the hull for one of the simulations. 

\subsection{Boundary and Initial Conditions}
\label{sec:BIC}
Standard Dirichlet boundary conditions appropriate to incompressible flow around a body are used for all simulations.
A velocity inlet boundary condition of 2 m/s was applied on the left-hand side of the domain, and a zero pressure boundary condition was applied on the right-hand side of the domain.
A no-slip boundary condition was applied to the hull surface, and the remaining surfaces (consisting of the side wall, symmetry wall, and far-field) were set to
symmetry. Simulations were performed at a flow speed of 2 m/s and the assumed inflow
turbulence intensity level was 4\%, assuming medium turbulence since it is a low-speed flow. Initial values for $k$ and $\omega$ were taken to be 0.01 $\text{m}^2/\text{s}^2$ and 57 $\text{s}^{-1}$, respectively.  Since we assumed the Newtonian model, which assumes kinematic viscosity to be constant, we assign a fixed kinematic viscosity of $1.7\; \text{mm}^2/\text{s}$ .
 
\subsection{Solver settings}
\label{subsec:SS}
OpenFOAM's solver includes various settings controlled by the \texttt{fvSchemes}, \texttt{controlDict}, and \texttt{fvSolution} files.  We report all the settings and parameter values used in our simulations, for sake of reproducibility.
We use default values unless otherwise specified.

The \texttt{fvSchemes} file controls variables related to the spatial finite volume discretization.  \texttt{gradSchemes} is set to Gauss linear, \texttt{laplacianSchemes} to Gauss linear uncorrected, and the \texttt{div(phi,k)} and \texttt{div(phi, omega)} terms set to bounded Gauss upwind.
The \texttt{controlDict} file controls simulation parameters like start time, end time, and time step. The maximum simulation time is set to 500 s or until convergence. The solver is set to \texttt{SimpleFoam}, which is a steady-state solver for incompressible, turbulent flow that uses the SIMPLE algorithm \cite{patankar1983calculation}.
In the \texttt{fvSolution} file, the pressure solver used was GAMG (geometric agglomerated algebraic multigrid preconditioner) with \texttt{symGaussSeidel} smoother with a tolerance of $1.0\times 10^{-7}$ and a relative tolerance of $0.01$. For $U$, $k$, and $\omega$, \texttt{smoothsolver} was used with the \texttt{GaussSeidel} smoother. The tolerance in each case is set at $1.0\times 10^{-7}$ with \texttt{relTol} equal to $0.1$.
For the potential flow, \texttt{nNonOrthogonalCorrectors} was set to $8$. 

\subsection{Simulation results}

\begin{figure}[ht!]
    \centering
    \subfloat[]{
        \includegraphics[width=0.49\textwidth]{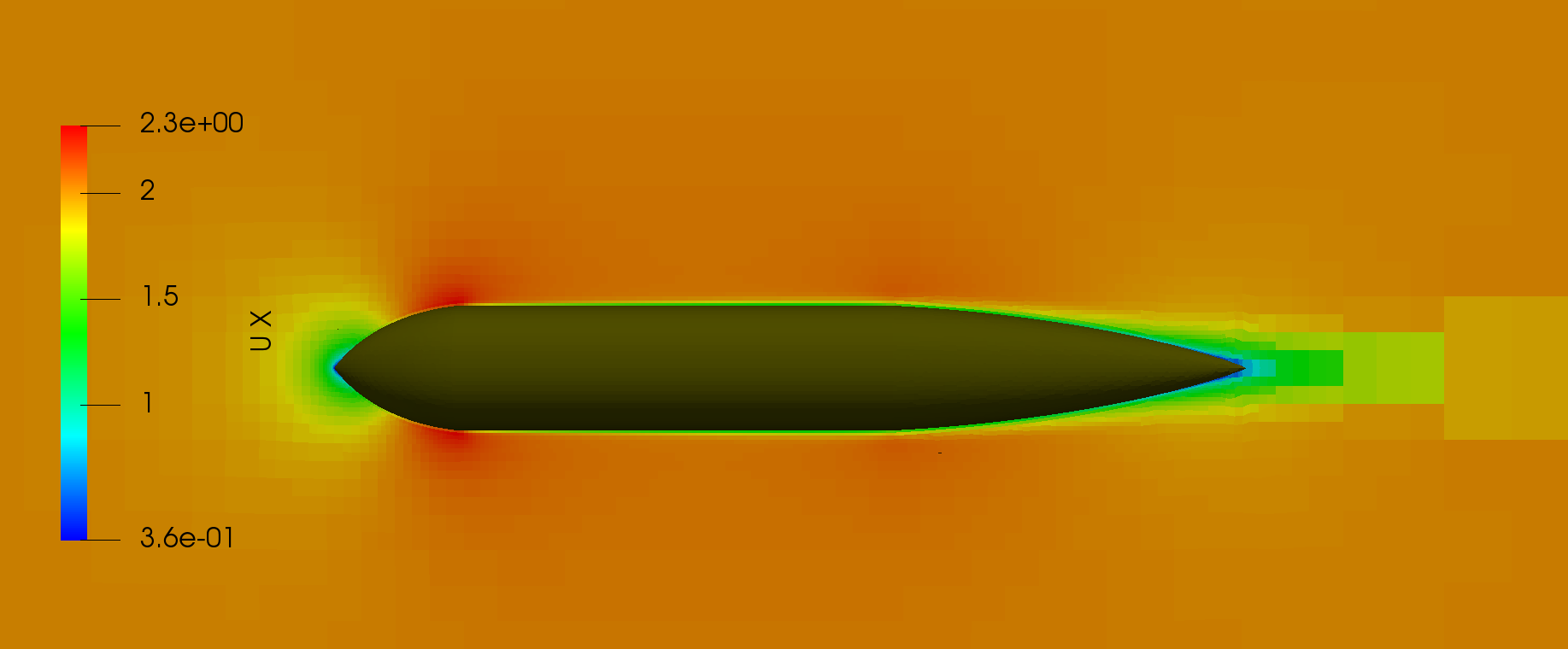}
        \label{fig:steadystateflowproperties:vf}
    }
    \subfloat[]{
        \includegraphics[width=0.425\textwidth]{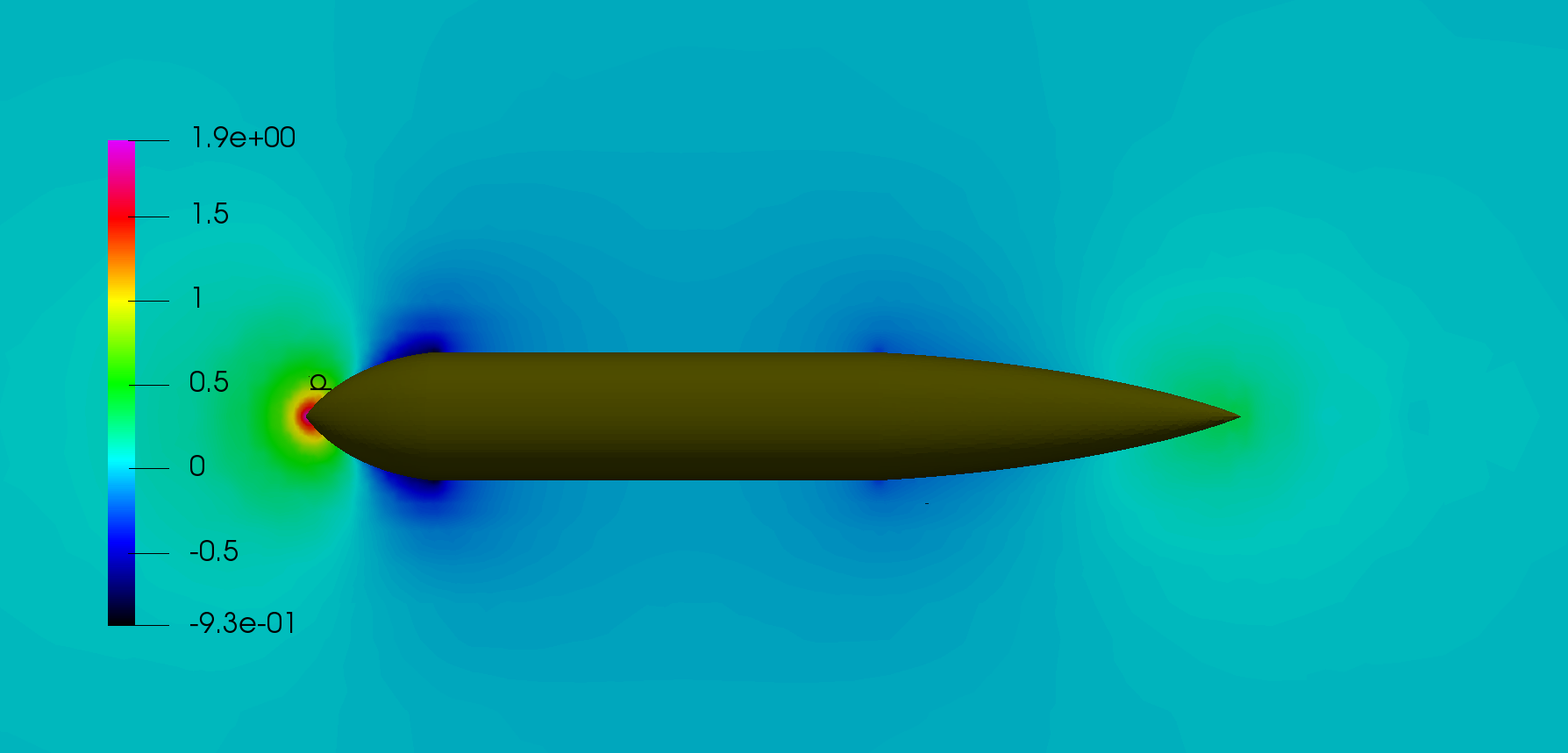}
        \label{fig:steadystateflowproperties:pf}
    }
    \caption{An example steady state flow, with the flow field colored by \protect\subref{fig:steadystateflowproperties:vf} mean axial velocity (meters/second), \protect\subref{fig:steadystateflowproperties:pf} pressure (Pascals).}
    \label{fig:steadystateflowproperties}
\end{figure}

Figure \ref{fig:steadystateflowproperties} shows the simulation outcome of one of the simulation experiments using a UUV hull design in OpenFOAM using the above-mentioned setting and methodology.
ParaView \cite{ayachit2015paraview} was used to create the contour plot of mean axial velocity and pressure field in the flow region.

\section{Direct optimization with CFD in the loop}
\label{sec:DO}

Direct optimization of a UUV hull design using CFD in the loop has been demonstrated in recent works \cite{gao2016hull}.
For design optimization, some commercial CFD simulation software offerings even come bundled with different design exploration methods, response surface methods, sensitivity analysis methods, as well as genetic algorithms and other traditional optimization algorithms.
Recent advances in AI-based optimization methods hold promise of better performance than existing optimization methods.
We are interested in comparing a diverse mix of algorithms to determine which has best sample efficiency and convergence behavior in the case of UUV hull design optimization.
To this end, we selected six different design of experiment (DOE) and optimization algorithms: 
\begin{enumerate}
    \item Monte Carlo with sampling from uniform random distribution 
    \item maximin Latin Hypercube-based optimization \cite{morris1995exploratory}
    \item Vanilla Genetic Algorithm \cite{holland1992adaptation, vose1999simple} 
    \item Nelder-Mead \cite{nelder1965simplex}
    \item Bayesian Optimization---Lower Confidence Bound (BO-LCB) \cite{clark1961greatest, kushner1964new,zhilinskas1975single,movckus1975bayesian,jones1998efficient}
    \item Bayesian Optimization---Expected Improvement (BO-EI) \cite{kushner1964new,zhilinskas1975single,movckus1975bayesian,srinivas2012information}
\end{enumerate}
Since none of these algorithms is novel, we do not dwell on their details here, but we provide reviews of each of the algorithms in \ref{sec:optimization-methods}.
We note that when it comes to optimization algorithms, there are many variants.  In this work, we attempted to select popular choices from several classes of gradient-free algorithms: traditional design of experiments (DOE), GA and its variants, and modern AI-based approaches. Within the first category, we chose two algorithms that have been used in engineering for decades, and within the second category, we chose algorithms that we feel represent modern workhorses for solving optimization problems.
Bayesian optimization (BO) is a powerful AI-based optimization algorithm, although its application in engineering domains is very nascent.
To our knowledge, there is no other work on UUV hull design that has implemented BO and evaluated its performance against other traditional methods.

For evaluation of the performance of these optimization algorithms, we are interested in two key metrics:

\begin{enumerate}
    \item \textbf{Sample efficiency}: The UUV design optimization problem involves running computationally expensive CFD simulations. This motivates the consideration of sample-efficient optimization, i.e., finding the optimal design with few samples. Sample efficiency is a relative way to measure the number of evaluations an optimization algorithm takes to find the optimal design when compared to other methods. Since there is some randomness in the optimization process as well, as it may depend on initial conditions and other specifics of the problem, we are interested in finding expected sample efficiency  $\mathbb{E}(|S|)$, where $S$ is the set of candidate designs an optimization algorithm considers and $|S|$ is the cardinality of $S$.
    \item \textbf{Convergence behavior}: Another aspect that we are interested in investigating is variation in the optimization process. By inspecting and comparing the differences in variation between the optimization processes, we can get a sense of convergence. Here variance gives us an empirical understanding of how much variability in the optimal design can happen if different initial designs are used in optimization.
\end{enumerate}

To formally define the optimization problem, we denote the 3D hull shape by $\Omega$, which is parameterized on a multivariate parameter $X$, i.e.\ $\Omega = \Omega(X)$.
Let $f:\Omega \mapsto F$ be a function that maps the 3D hull shape design $\Omega$ to the steady-state drag force $F$ experienced by the hull as it travels through the fluid.
Our optimization goal is then to minimize $f$ on the domain of interest $DS$, $X \subset DS$:
\begin{equation}
     \Omega^{*} =\underset{X \in DS }{\mathrm{argmin}} f(X)
\end{equation}

As a real-world case study, we consider the Remus100 class design \cite{winey2020modifiable} as the basis for our design space.
The Remus100 design is a Myring hull-based shape whose diameter and total length are $0.191\text{m}$ and $1.33\text{m}$, respectively.
We assume both diameter and total length as constants for purposes of optimization, while all other Myring hull shape parameters ($a,b,c,n,\theta$) are variable and part of the optimization routine.  The total length and diameter of the vehicle are fixed to 1.33 m and 0.191 m, similar to Remus100. The search domain for design parameters ($a,b,c,n,\theta$) is given in Table \ref{tab:ds_single}.

\begin{figure}[ht!]
    \centering
    \includegraphics[trim={3cm 13.5cm 3cm 3.1cm},clip]{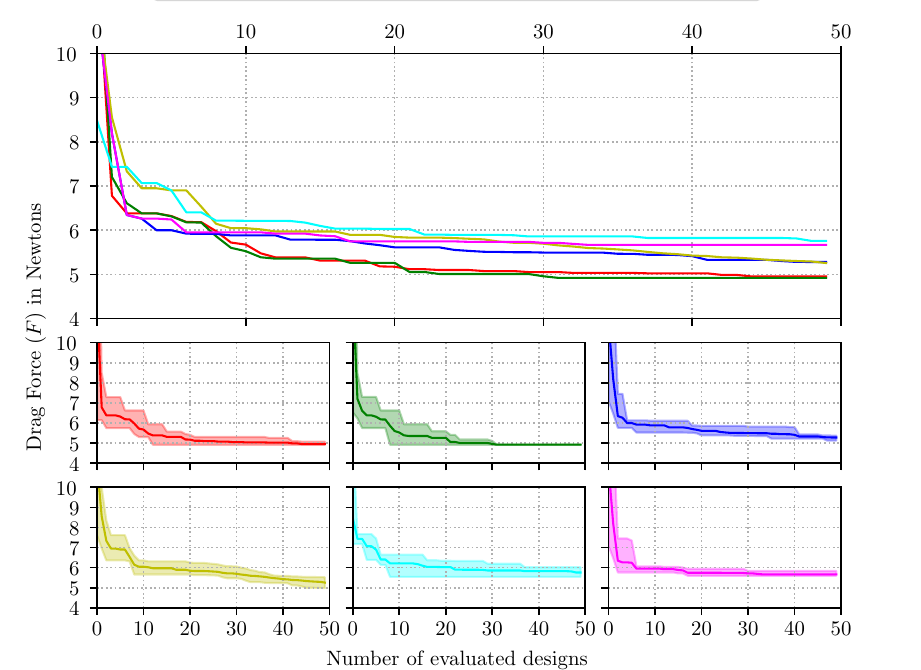}
    \caption{Optimal $F$  vs.\ the number of evaluated designs. The top plot shows the expected optimal drag found by each optimization algorithm as the design space is explored. The expectation is calculated over five different optimization runs. The bottom six plots show the mean (the thick line) and the variance (shaded region) of drag forces found by using different optimization algorithms. Here BO-EI refers to Bayesian Optimization---Expected Improvement, BO-LCB refers to Bayesian Optimization---Lower Confidence Bound, GA refers to Genetic Algorithm, LHC refers to maximin Latin Hypercube, VMC refers to Vanilla Monte Carlo, and NM refers to Nelder-Mead method.}
    \label{fig:opt_traces}
\end{figure}

\begin{table}
\centering
\normalsize
\captionsetup{justification=centering,format=plain,font=small, labelfont=bf}
\begin{tabular}{lccc} 
\toprule
\textbf{Parameter} & \hfil \textbf{Symbol} & \hfil \textbf{ Minimum } & \hfil \textbf{ Maximum } \\
\midrule
Length of nose section & $a$ & $50$ mm & $573$ mm  \\
Length of tail section  & $c$ & $50$ mm & $573$ mm \\
Index of nose shape & $n$ & $1.0$ & $50.0$ \\
Tail semi-angle  & $\theta$ & $0^o$ & $50^o$  \\
$b=1910 -(a+c)$  \\
\bottomrule
\end{tabular}
\vspace{0mm}
\caption{Design Space: Range of design space parameters for optimization.}
\label{tab:ds_single}
\end{table}
The budget for optimization is set to 50 evaluations. Figure \ref{fig:opt_traces} shows the outcome of the design and its drag force using each of the algorithms we evaluated. It can be seen from the top plot that BO-LCB converges to the most optimal design observed in the fewest number of samples.
BO-EI is comparable in expected sample efficiency to BO-LCB, but the variance in finding the optimal design is much higher than BO-LCB (refer to the lower plots of Figure \ref{fig:opt_traces}); BO-LCB always finds the same optimal design after about 30 iterations. Other methods find strictly worse optimal designs within the given budget.

\begin{figure}[ht!]
    \centering
     \begin{mdframed}[backgroundcolor=aquamarine,topline=false,bottomline=false,rightline=false,leftline=false]
    \begin{tabular}{|c|c|}
        \hline
        \hypertarget{fig:optsamples:BOEI}{}
        \hypertarget{fig:optsamples:BOLCB}{}
        
        & \\
          \includegraphics[width=0.45\textwidth]{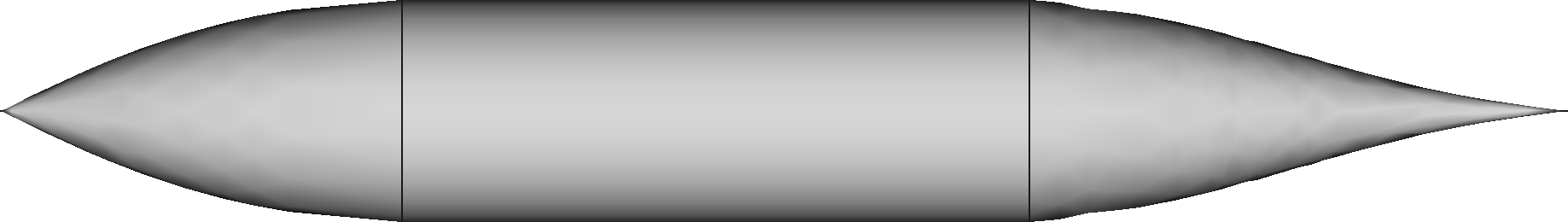} & 
          \includegraphics[width=0.45\textwidth]{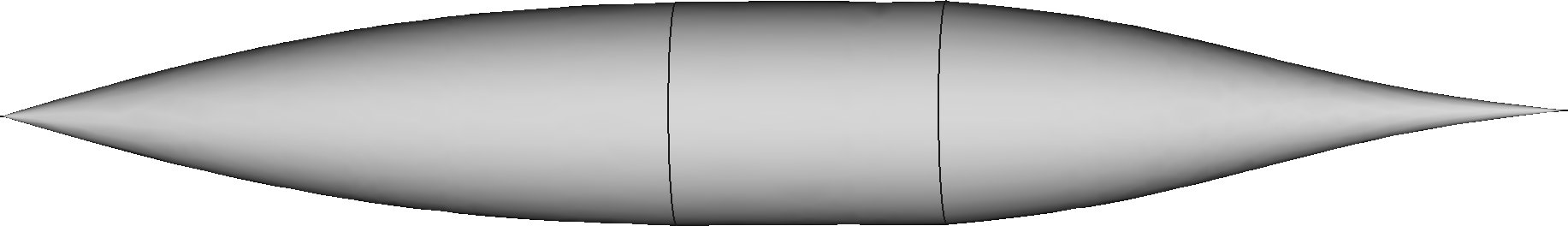}\\
          (a) &  (b)   \\
\hline
            \hypertarget{fig:optsamples:GA}{}
            \hypertarget{fig:optsamples:NM}{}
          & \\
            \includegraphics[width=0.45\textwidth]{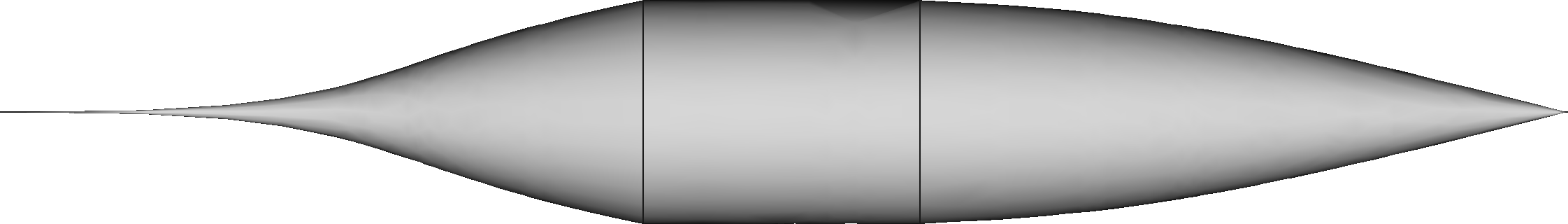} &
            \includegraphics[width=0.45\textwidth]{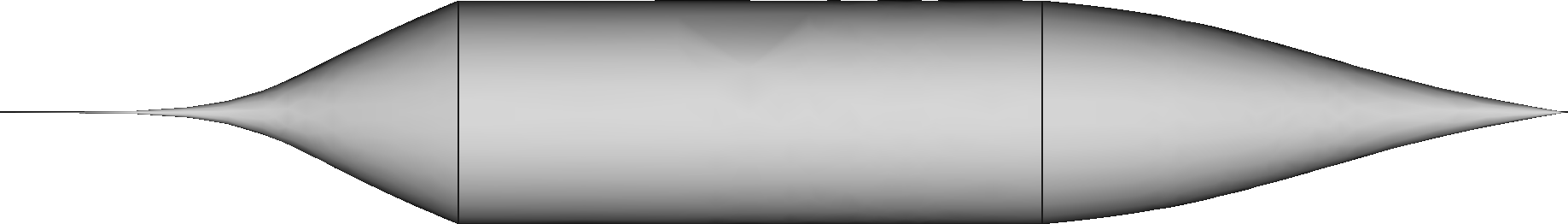} 
            \\
          (c) &  (d) \\
\hline
        \hypertarget{fig:optsamples:maximin}{}
        \hypertarget{fig:optsamples:VMC}{}
        & \\
         \includegraphics[width=0.45\textwidth]{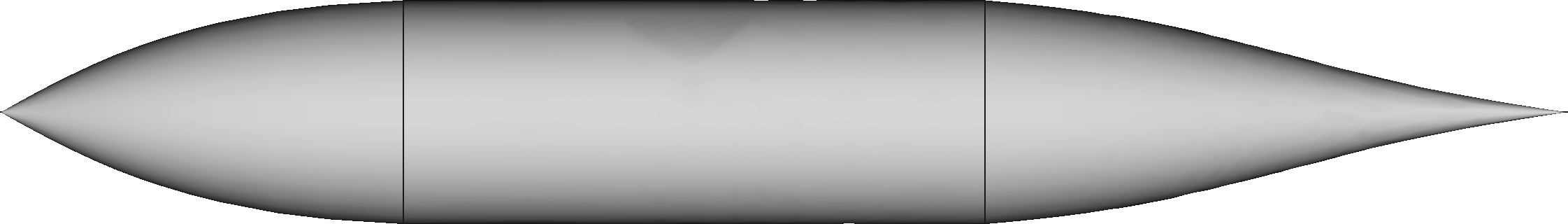}
        & \includegraphics[width=0.45\textwidth]{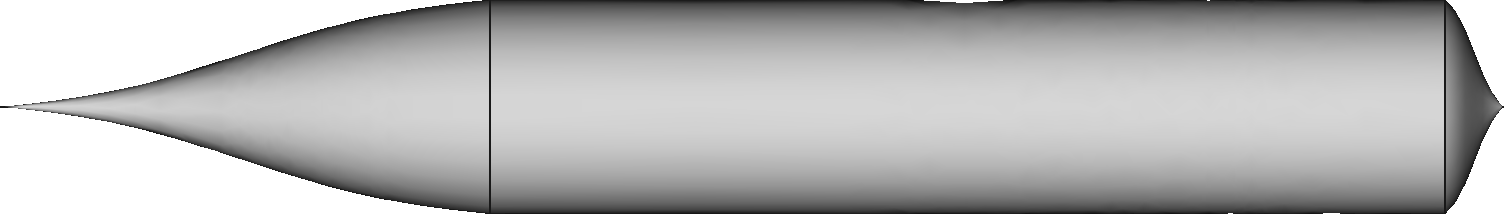} \\
        (e) &  (f) \\
\hline
    \end{tabular}
    \end{mdframed}
    \caption{Results of one optimization run: Optimal design discovered after exhausting the budget of simulation, using different optimization algorithms. \protect\hyperlink{fig:optsamples:BOEI}{(a)} Bayesian optimisation---Expected Improvement, \protect\hyperlink{fig:optsamples:BOLCB}{(b)} Bayesian optimization---Lowest confidence bound, \protect\hyperlink{fig:optsamples:GA}{(c)} Genetic Algorithm, \protect\hyperlink{fig:optsamples:NM}{(d)} Nelder Mead, \protect\hyperlink{fig:optsamples:maximin}{(e)} maximin Latin Hypercube and \protect\hyperlink{fig:optsamples:VMC}{(f)} Vanilla Monte Carlo. The design produced by BO-LCB (design \protect\hyperlink{fig:optsamples:BOLCB}{(b)}) has lowest drag.}
    \label{fig:optsamples}
\end{figure}

Figure \ref{fig:optsamples} shows the results of one of the five trials: the optimal UUV hull shape discovered by all methods after 50 samples/iterations.
It can be seen that the design obtained by BO-LCB is a highly streamlined body.
The design obtained from BO-EI is the closest to the optimal design obtained from the BO-LCB. The designs obtained from GA and Nelder-Mead have degenerate nose and tail features and are not close to the near-optimal design successfully identified by BO-LCB.
The design obtained by maximin LHC is better (similar to that found by BO-EI), yet still far from the optimal design. \section{Surrogate-Based Optimization}
\label{sec:ga_sbo}

Surrogate-based optimization refers to the idea of accelerating the outer loop of an optimization process by using surrogates for the objective and constraint functions.
Surrogates also allow for the optimization of problems with non-smooth or noisy responses and can provide insight into the nature of the design space.
In our application, UUV hull design optimization, replacing computationally-expensive CFD simulations with efficient surrogates promises significant speedups.

 \subsection{Dataset Generation}
In order to construct surrogate models, we build a dataset of mappings between hull shape and steady-state drag force.  OpenFOAM simulation output is taken as ground truth for this purpose, though any other simulation tools and algorithms could be used.
We consider the space of solutions with a range shown in Table \ref{tab:ds}; our aim was to cover the design space for small UUV shapes.
The design of the hull is randomly sampled from this design space.

The boundary and initial conditions and the solver setting are kept the same across all experiments as explained in Sections \ref{sec:BIC} and \ref{subsec:SS}, respectively.
Each simulation took an average of 10 minutes on a 16-core CPU.  (For all our experiments, we used a system with an Intel i9-9820X 10-core CPU @ 3.30GHz and 62GB of RAM, and used 8 out of available 10 cores for running OpenFOAM.)
The number of the generated designs was 3,333, which includes $312$ invalid designs that were generated and then discarded, leaving 3,021 valid designs. 
We attempted using smaller data sets, e.g.\ of size $500$, but found that training and testing accuracy of the surrogate models was undesirable.
We expect that further accuracy would be obtained with larger data sets but were conscious of computational budgets.

\begin{table}
\centering
\normalsize
\captionsetup{justification=centering,format=plain,font=small, labelfont=bf}
\begin{tabular}{lccc} 
\toprule
\textbf{Parameter} & \hfil \textbf{Symbol} & \hfil \textbf{ Minimum } & \hfil \textbf{ Maximum } \\
\midrule
Diameter of middle section & D & $100$ mm  & $400$ mm \\
Length of nose section & $a$ & $50$ mm & $600$ mm  \\
Length of middle section & $b$ & $1$ mm & $1850$ mm  \\
Length of tail section  & $c$ & $50$ mm & $600$ mm \\
Index of nose shape & $n$ & $1.0$ & $5.0$ \\
Tail semi-angle  & $\theta$ & $0^o$ & $50^o$  \\
Total length & $l=a+b+c$ &  &  \\
\bottomrule
\end{tabular}
\vspace{0mm}
\caption{Design Space: Range of design parameters for surrogate modeling.}
\label{tab:ds}
\end{table}

 \subsection{Neural Architecture}
 Our surrogate model is created by training a neural network.
 Our neural network model is a custom architecture designed based on a fully connected network, which is widely used for a wide variety of tasks.
Our network architecture consists of 9 layers. The main components in the network are: the fully connected nonlinear layer, dropout layer, skip connections, and output layer.
 The inputs are stacked in the following order: $x_1=a$, $x_2=b$, $x_3=c$, $x_4=d$, $x_5=n$, $x_6=\theta$, and the concatenated vector $X=[x_1,x_2,x_3,x_4,x_5,x_6]$ of input data is normalized across each feature using min-max normalization:
  $$\hat{X}[:,i]= \dfrac{X[:,i]-\min(X[:,i])}{\max(X[:,i]-\min(X[:,i])} .$$
  The normalized vector $\hat{X}$ is fed into the neural network as input for prediction. 
  The fully connected layer encompasses one linear matrix operation with a trainable weight parameter ($W$) that is then added to a trainable bias ($b$).
  The outcome of this linear operation is passed through a nonlinear ReLU (Rectified Linear Unit \cite{xu2015empirical}) activation function, defined by $f(x)=\max(X,0)$. Each section of the fully connected layer computes $\max(WX+b,0)$ and passes this information to the next layer.
  Two dropout layers were added in between these fully connected layers for regularization and to prevent overfitting and improve generalization.
  The dropout layer computes a Hadamard product between the last layer's output vector and a generated random Bernoulli variable $z$ of the size equal to last layer's output vector, where ($z_i \sim B(p)$ and $z=\{z_1, z_2,...,z_n\}$).  The dropout parameter $p$ decides the probability of Bernoulli's output $B\in \{0,1\}$.   Here, $p$ is the probability of outcome $1$, and $(1-p)$ is the probability of outcome $0$.
  The combined output of the fully connected layer and dropout layer can be written as $\max(WX+b,0) \otimes z$, where $\otimes$ is the Hadamard product.
   The number of parameters in our architecture is 22,249.
   
The dropout layer and cross-validation-based early stopping are used to counter overfitting in our regression model.
   The dropout parameter in our network is set to $0.2$ for both dropout layers.   
  We also added a skip connection / residual layer at the inner hidden layer, which facilitates smooth learning and handles vanishing or exploding gradients \cite{he2016deep}.
  Our output layer is a linear layer, with only one linear unit. 
  
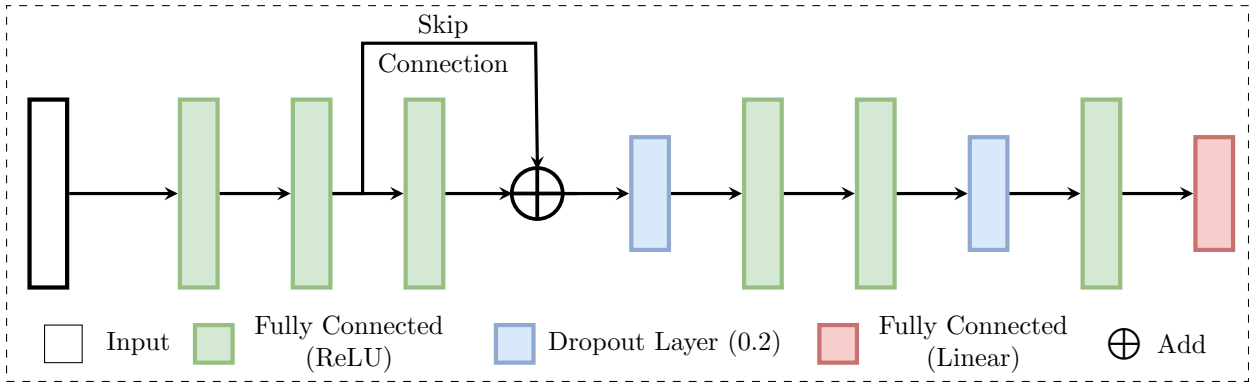
\begin{figure}[ht!]
        \centering
        \captionsetup{justification=centering}
        \begin{tikzpicture}
\tikzstyle{fc} = [ultra thick, rectangle, minimum width=0.5cm, minimum height=2.5cm, text centered, draw=goodGreenBorder, fill=goodGreen, align=center];
\tikzstyle{dropout} = [ultra thick, rectangle, minimum width=0.5cm, minimum height=1.5cm, text centered, draw=goodBlueBorder, fill=goodBlue, align=center];

\tikzstyle{lfc} = [ultra thick, rectangle, minimum width=0.5cm, minimum height=1.5cm, text centered, draw=goodPinkBorder, fill=goodPink, align=center];
    \node(input) [fc, draw=black, text centered, align=center, fill=none] at (0,0) {};
    \node(skip connection) [rectangle, draw=none, fill=none, right of=input, xshift=4.25cm, yshift=2cm, text centered, align=center]{Skip\\[0.3em]Connection};
    
    \node(fc1)[fc, right of=input, xshift=1cm]{};
    
    \node(fc2)[fc, right of=input, xshift=2.5cm]{};
    
    \node(fc3)[fc, right of=input, xshift=4cm]{};
    
    \node(concatenate) [rectangle, right of=input, draw=none, fill=none, xshift=5.5cm]{\huge $\bigoplus$};
    
    \node(drop1)[dropout, right of=input, xshift=7cm]{};
    
    \node(fc4)[fc, right of=input, xshift=8.5cm]{};      \node(fc5)[fc, right of=input, xshift=10cm]{};
    \node(drop2)[dropout, right of=input, xshift=11.5cm]{};
    \node(fc6)[fc, right of=input, xshift=13cm]{};
    \node(fc7)[lfc, right of=input, xshift=14.5cm]{};

    \draw[very thick, -stealth] (input) -- (fc1);
    
    \draw[very thick, -stealth] (fc1) -- (fc2);
    
    \draw[very thick, -stealth] (fc2) -- (fc3);
    
    \draw[very thick, -stealth] (fc3) -- (concatenate.west) - ++(0.2, 0);
    
    \draw[very thick, -stealth] (concatenate.east) -- ++(-0.18, 0) -- (drop1);
    
    \draw[very thick, -stealth](fc2.east) -- ++(0.4, 0) -- ++(0, 2) -- ++(2.3, 0)  -- (concatenate.north) -- ++(0, -0.17);
    
    \draw[very thick, -stealth] (drop1) -- (fc4);

    \draw[very thick, -stealth] (fc4) -- (fc5);
    \draw[very thick, -stealth] (fc5) -- (drop2);
    \draw[very thick, -stealth] (drop2) -- (fc6);
    
    \draw[very thick, -stealth] (fc6) -- (fc7);

\begin{scope}[xshift=7.7cm]
\node(bbox) [rectangle, fill=none, draw=black, dashed, minimum height=5cm, minimum width=\textwidth] at (0, 0) {};
    
    \node(legend) [rectangle, fill=none, draw=none, below of=bbox, yshift=0cm]{};
    
    \node(legendinput)[below of=legend, rectangle, fill=none, draw=black, minimum height=0.5cm, minimum width=0.5cm, text centered, xshift=-7.5cm]{};
    
     \node(textinput1) [rectangle, fill=none, draw=none, right of=legendinput, yshift=0cm, text centered, align=center]{Input};

    \node(legendfc)[fc, below of=legend, minimum height=0.5cm, minimum width=0.5cm, xshift=-5.5cm]{};
    
     \node(textinput2) [rectangle, fill=none, draw=none, right of=legendfc, yshift=0cm, xshift=0.8cm, text centered, align=center]{Fully Connected \\ (ReLU)};
    
     \node(legenddrop)[dropout, below of=legend, minimum height=0.5cm, minimum width=0.5cm, xshift=-1.5cm]{};
     
      \node(textinput3) [rectangle, fill=none, draw=none, right of=legenddrop, yshift=0cm, xshift=1cm, text centered, align=center]{Dropout Layer (0.2)};
     
     \node(legendlfc)[lfc, below of=legend, minimum height=0.5cm, minimum width=0.5cm, xshift=2.8cm]{};
     
     \node(textinput4) [rectangle, fill=none, draw=none, right of=legendlfc, yshift=0cm, xshift=0.8cm, text centered, align=center]{Fully Connected \\ (Linear)};
     
     \node(legendadd)[rectangle, below of=legend, minimum height=0.5cm, minimum width=0.5cm, xshift=6.6cm]{\large$\bigoplus$};
     
     \node(textinput4) [rectangle, fill=none, draw=none, right of=legendadd, yshift=0cm, xshift=-0.23cm, text centered, align=center]{Add};
\end{scope}

\end{tikzpicture}         \caption{Our DNN surrogate architecture.  The network is trained to map design inputs ($a,b,c,n,\theta$) to predicted steady-state drag forces $F$.}
        \label{fig:DNN_cfd}
\end{figure}
 
\subsection{Training Details}
\begin{figure}[ht!]
    \centering
    \includegraphics[width=0.8\textwidth]{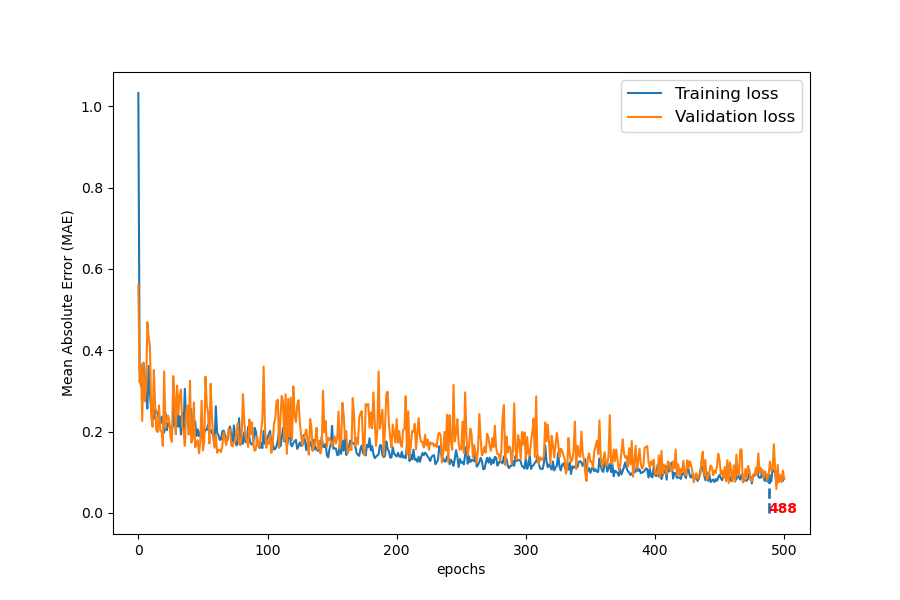}
    \caption{The average training and validation loss vs.\ epoch number during the training process.  We select the model parameters at epoch 467, out of the maximum 500 epochs considered.}
    \label{fig:train_loss}
 \end{figure}

The neural network is trained using a back-propagation algorithm, with the evaluated cost function based on the error between the predicted output of the trained network and the ground truth generated from CAD--CFD simulation.
For training our neural network model, we used the Adam \cite{kingma2014adam} optimizer, built into PyTorch. We also tuned the learning rate, loss function and training epoch to find the best-trained model with minimum training and validation error. By experimentation, we found learning rate of $0.0003$ worked best. Initially, we used mean-squared error ($L_2$ loss) to train the network, but we found that $L_1$ loss resulted in a lower mean prediction error. The other hyperparameters of Adam were kept as default values used in its original implementation \cite{kingma2014adam}. The weight parameter is initialized using Xavier-normal \cite{glorot2010understanding} with constant zero bias.
In addition to the dropout layer used in the network architecture, we also deployed early stopping to prevent overfitting and to improve generalization on our model on training data. 
For training and testing purposes, we first split the total collected data points (3,021) into two sets: the first set consists of 2,400 datapoints, and the second set consists of $621$.
For early stopping and cross-validation of the model, we divide the first set of data (2,400 datapoints) further, where $90\%$ of the data is used for training to minimize the cost error function and $10\%$ of the data is used for validation and early stopping of training to get better generalization error from the trained network.
The validation error is compared to the training error after each epoch of training, and during the training process, we only save the updated model when the gap between the training error and validation error is reduced. The set of $621$ datapoints is used for testing the performance of the trained surrogate. Figure \ref{fig:train_loss} shows how training and validation loss evolve as the network is trained.

\subsection{Surrogate Performance}
To test the prediction accuracy and generalization capability of the trained surrogate, we used the following metrics: 
\begin{enumerate}
    \item \textbf{Residual}, $\Delta Z= \tfrac{(F^{gt}- F^p)}{F^{gt}}$  
    
    \item \textbf{Accuracy} ($\alpha$), percentage of test samples whose residual is within acceptable error (we choose 5\%, i.e., $|\Delta Z| \leq 0.05$)
    \item \textbf{Mean Absolute Percentage Error}, $MAPE= \frac{1}{n}\sum_{i=1}^{n}\Big|\tfrac{(F_{i}^{gt}- F_{i}^p)}{F_{i}^{gt}}\Big|$ 
    \item \textbf{Outliers} ($\eta$), percentage of data samples on which trained surrogate prediction error is more than acceptable range, $|\Delta Z| >0.10$.
\end{enumerate}
Here, $F^{gt}$ is the ground truth drag force estimated by running CFD simulations, and $F^p$ is the predicted drag force from the trained DNN surrogate model. Generally, $MAPE$ is the loss function and is also used to report the accuracy of a regression model. $MAPE$ estimates the average prediction accuracy and fails to provide an explanation of whether the error is due to some outliers or generalization error. We define outliers as  data points whose predicted values are more than $10\%$ away from the ground truth. We measure the number of outliers as $\eta$. We also report another metric, $\alpha$, which is the percentage of test samples whose residual is within acceptable error ($\pm 5\%$). The choice of ``acceptable'' error bounds (e.g., $\pm 10\%$ for $\eta$ and $\pm 5\%$ for $\alpha$) is up to the user and depends on experience in the domain.

These metrics are reported in Table \ref{tab:space} for the training and validation sets, and most importantly, for the test set.
Our surrogate model makes accurate predictions in the vast majority of test cases, and among the inaccurate predictions, only 14 of the 621 test data points are reported as outliers. Similar prediction accuracy is observed for the validation and training data, which indicates successful training.

\begin{figure}[!ht]
    \centering
    \includegraphics[trim={0.1cm 0.1cm 0.1cm 0.1cm},clip]{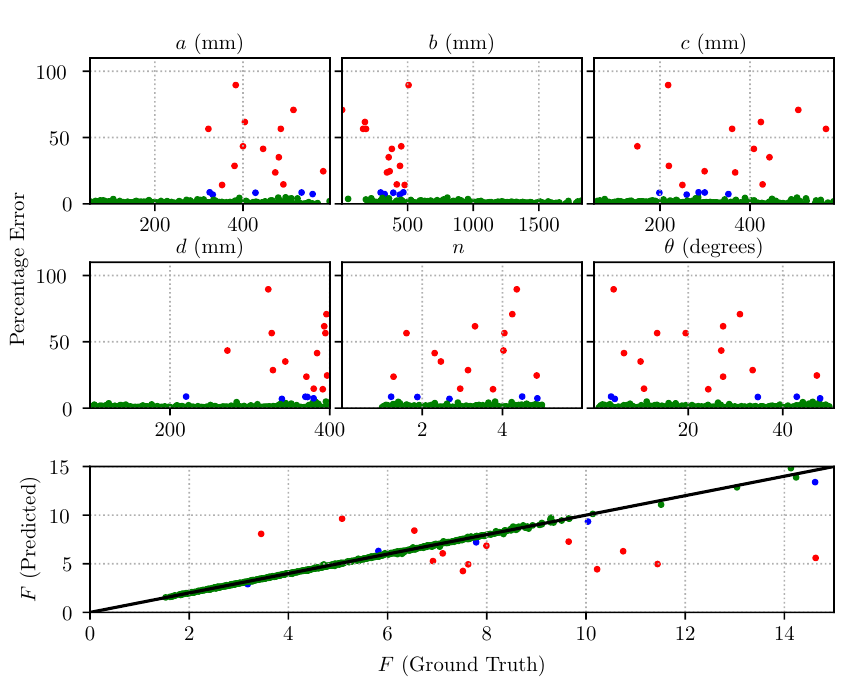}
  \caption{(Upper rows) Percentage error for test data as a function of the various design parameters. (Bottom row) Predicted values (via DNN surrogate) vs.\ ground truth steady-state drag forces $F$. In all subfigures, green markers denote predictions with less than 5\% error; blue markers denote predictions between 5\%--10\% error; red markers denote predictions with errors greater than 10\%.  14 of 621 test points are outliers (red).}
    \label{fig:residuals-vs-params}
\end{figure}

Figure \ref{fig:residuals-vs-params} visualizes all the test data points, including outliers.
The figure helps illuminate where additional sampling could improve the performance of our network.
For example, supplying more training data with $b$ below 500 mm, or $d$ closer to 400 mm, would likely improve network performance in those regimes (where there are currently disproportionate numbers of outliers).

 \begin{table}
\captionsetup{justification=centering,labelsep=period}
\caption{Performance of our DNN surrogate for predicting steady-state drag force on UUV hull designs.}
\label{tab:space}
\centering
\begin{tabular}{cccccc} 
\toprule
\textbf{Data Set} & \textbf{Data Points} & \textbf{Accuracy ($\alpha$)} &\textbf{MAPE} &\textbf{ Outliers ($\eta$)} & \textbf{\% Outliers}\\
\midrule
Training & 2160  &  97.69\% & 1.12\% & 29  & 1.34\%\\
Validation & 240 & 98.33\% & 1.36\% & 3 & 1.25\%\\
Test & 621 & 96.94\% & 1.85\% & 14 & 2.25\%\\
\bottomrule
\end{tabular}
\end{table}

\subsection{Hull Optimization Results using Trained Surrogate}
\label{sec:surrogate-results}
Our trained neural network-based surrogate can characterize the input-output relationship between design inputs and drag.
Since the surrogate achieves an acceptable accuracy---but at a far lower computational cost than RANS CFD simulations---we consider using this high-speed data-fit low-fidelity surrogate model in the outer loop of our UUV hull shape optimization procedure, rather than using OpenFOAM.
When a cheap-to-evaluate surrogate is available, Genetic Algorithm is the popular choice of optimization algorithm \cite{bradford2018efficient}.

\begin{figure}[ht!]
    \centering   
    \includegraphics[width=0.94\textwidth,trim={2.5cm 11.5cm 1cm 2.1cm},clip]{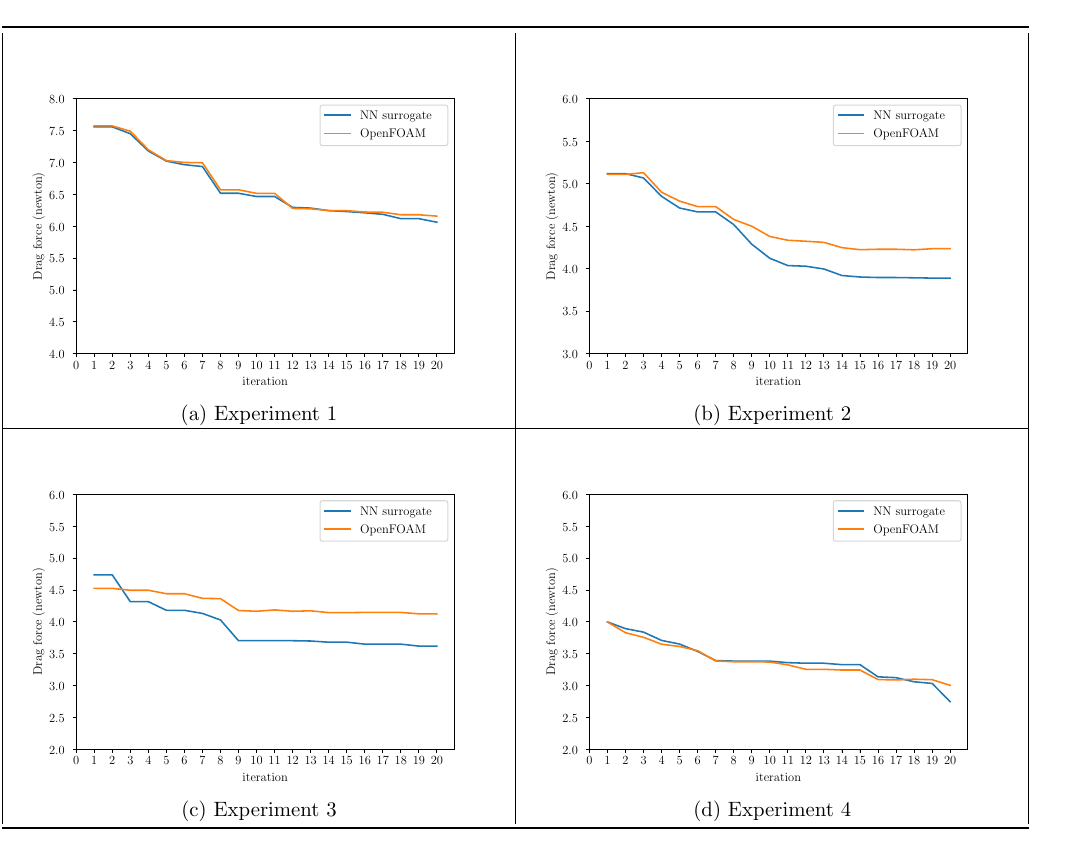}
   \caption{Results of the optimization runs: Drag of best design obtained during each iteration using the surrogate in the loop with GA and drag estimation by the OpenFOAM by running full CFD analysis. 
   The four experiments use different constraints on the design parameters, listed in Section \ref{sec:surrogate-results}.}
    \label{fig:ga_nn_foam}
\end{figure}

To evaluate the benefits of using our surrogate model in the design optimization loop, we evaluate (1) the computational speedup, and (2) whether the predicted values still generally agree with the results obtained when using OpenFOAM in the optimization loop.
Given that one of our OpenFOAM simulations can take 10--20 minutes, and that evaluating our neural network surrogate takes around $0.002 \; \text{s}$, we expect a speedup of around one hundred thousand times.
To compare the prediction quality of our surrogate during design optimization, we evaluate it in four different experiments, each with different requirements/constraints: 
\begin{itemize}[nosep]
    \item Experiment 1: Diameter: 180 mm; Total length: 1750 mm
    \item Experiment 2: Diameter: 150 mm; Total length: 1500 mm
    \item Experiment 3: Diameter: 165 mm; Total length: 2000 mm
    \item Experiment 4: Diameter: 120 mm; Total length: 1250 mm
\end{itemize}
During optimization, we run GA in the loop with our trained surrogate, and at the end of each iteration of GA, we compare the drag value of the best-found design by GA (using the surrogate model) against the value computed (for the same design) using the full CFD simulation (RANS + $\kappa$-$\omega$ SST). We do this comparison after every iteration and report the drag value estimated by the surrogate and by the OpenFOAM simulation.  

Figure \ref{fig:ga_nn_foam} shows the results of these four experiments. It can be observed from the plots that in some cases, there is a strong correspondence between the predicted drag from the surrogate and from OpenFOAM throughout the entire optimization process.
In other cases, even though there is an error in the predicted drag between the surrogate and OpenFOAM, the trends in the functions over the course of the optimization loop are almost identical: the surrogate still correctly identifies which designs have relatively higher or lower drag.
Hence, in these cases, one could run OpenFOAM on the final design suggested by the surrogate-in-the-loop process in order to get an accurate drag value, and still be left with a near-optimal design.
Thus, surrogate-in-the-loop optimization succeeds in finding near-optimal design parameters with no reliance on OpenFOAM, and when the drag force value is required, at most one run of OpenFOAM can suffice.

 \section{Conclusions}
\label{sec:conclusion}
In this work, we conducted an empirical evaluation of different gradient-free optimization methods on a real-world UUV hull design problem.
Our empirical evaluation suggests that Bayesian optimization---Lower Confidence Bound (BO-LCB) has the best convergence behavior and sample efficiency in comparison to other selected gradient-free optimization methods.
We also presented a deep learning-based model for inference of drag force estimation on UUV shapes in a given design space using CFD simulations.
The trained deep learning-based surrogate model has high accuracy on test design parameters.
By using this trained deep learning-based surrogate, we demonstrated that the UUV shape optimization results in very close agreement with traditional CFD-in-the-loop design optimization. 
Notably, our deep neural network-based surrogate can solve design optimization problems four orders of magnitude faster than using RANS based fluid simulations.
Of course, a drawback of the surrogate model approach is the relatively large number of simulations required to train an accurate surrogate, but we emphasize that a surrogate model that has sufficient generalization capabilities does not need to be retrained, and practitioners could simply use such a pretrained model off-the-shelf without any concerns about training time.
We believe that both approaches considered in this work---sample-efficient optimization and surrogate modeling---have advantages and drawbacks, and that both have interesting directions for further development.

There are several avenues for extending the present work.
For instance, although we focused on small UUVs, it would be interesting to see if one surrogate model could be trained to handle all classes of UUVs while maintaining similar accuracy.
It would also be interesting to allow our parametric CAD model to support more complex features, such as dimples, which are known to have attractive behavior with respect to drag \cite{choi2006mechanism}.
One could also consider more complex flow conditions, such as higher turbulence or particulate flow.
Finally, although the neural network we used for our surrogate model was quite efficient, it would be interesting to perform an ablation study on parameters of the network architecture to better understand tradeoffs between parameters like the number of layers and the performance of the network. 

\section*{Acknowledgements}

This work is supported by DARPA through contract number FA8750-20-C-0537. Any opinions, findings, conclusions, or recommendations expressed are those of the authors and do not necessarily reflect the views of the sponsor.

\bibliographystyle{elsarticle-num-names}

\begin{thebibliography}{78}
\expandafter\ifx\csname natexlab\endcsname\relax\def\natexlab#1{#1}\fi
\providecommand{\url}[1]{\texttt{#1}}
\providecommand{\href}[2]{#2}
\providecommand{\path}[1]{#1}
\providecommand{\DOIprefix}{doi:}
\providecommand{\ArXivprefix}{arXiv:}
\providecommand{\URLprefix}{URL: }
\providecommand{\Pubmedprefix}{pmid:}
\providecommand{\doi}[1]{\href{http://dx.doi.org/#1}{\path{#1}}}
\providecommand{\Pubmed}[1]{\href{pmid:#1}{\path{#1}}}
\providecommand{\bibinfo}[2]{#2}
\ifx\xfnm\relax \def\xfnm[#1]{\unskip,\space#1}\fi
\bibitem[{Wang et~al.(2007)Wang, Lim, Khoo, and Wang}]{WANG2007395}
\bibinfo{author}{S.~Wang}, \bibinfo{author}{K.~Lim}, \bibinfo{author}{B.~Khoo},
  \bibinfo{author}{M.~Wang},
\newblock \bibinfo{title}{An extended level set method for shape and topology
  optimization},
\newblock \bibinfo{journal}{Journal of Computational Physics}
  \bibinfo{volume}{221} (\bibinfo{year}{2007}) \bibinfo{pages}{395--421}.
  \URLprefix
  \url{https://www.sciencedirect.com/science/article/pii/S0021999106002968}.
  \DOIprefix\doi{https://doi.org/10.1016/j.jcp.2006.06.029}.
\bibitem[{Hazra et~al.(2005)Hazra, Schulz, Brezillon, and Gauger}]{HAZRA200546}
\bibinfo{author}{S.~Hazra}, \bibinfo{author}{V.~Schulz},
  \bibinfo{author}{J.~Brezillon}, \bibinfo{author}{N.~Gauger},
\newblock \bibinfo{title}{Aerodynamic shape optimization using simultaneous
  pseudo-timestepping},
\newblock \bibinfo{journal}{Journal of Computational Physics}
  \bibinfo{volume}{204} (\bibinfo{year}{2005}) \bibinfo{pages}{46--64}.
  \URLprefix
  \url{https://www.sciencedirect.com/science/article/pii/S0021999104004061}.
  \DOIprefix\doi{https://doi.org/10.1016/j.jcp.2004.10.007}.
\bibitem[{Viquerat et~al.(2021)Viquerat, Rabault, Kuhnle, Ghraieb, Larcher, and
  Hachem}]{VIQUERAT2021110080}
\bibinfo{author}{J.~Viquerat}, \bibinfo{author}{J.~Rabault},
  \bibinfo{author}{A.~Kuhnle}, \bibinfo{author}{H.~Ghraieb},
  \bibinfo{author}{A.~Larcher}, \bibinfo{author}{E.~Hachem},
\newblock \bibinfo{title}{Direct shape optimization through deep reinforcement
  learning},
\newblock \bibinfo{journal}{Journal of Computational Physics}
  \bibinfo{volume}{428} (\bibinfo{year}{2021}) \bibinfo{pages}{110080}.
  \URLprefix
  \url{https://www.sciencedirect.com/science/article/pii/S0021999120308548}.
  \DOIprefix\doi{https://doi.org/10.1016/j.jcp.2020.110080}.
\bibitem[{Alexandrov and Santosa(2005)}]{ALEXANDROV2005121}
\bibinfo{author}{O.~Alexandrov}, \bibinfo{author}{F.~Santosa},
\newblock \bibinfo{title}{A topology-preserving level set method for shape
  optimization},
\newblock \bibinfo{journal}{Journal of Computational Physics}
  \bibinfo{volume}{204} (\bibinfo{year}{2005}) \bibinfo{pages}{121--130}.
  \URLprefix
  \url{https://www.sciencedirect.com/science/article/pii/S0021999104004103}.
  \DOIprefix\doi{https://doi.org/10.1016/j.jcp.2004.10.005}.
\bibitem[{Amstutz and Andrä(2006)}]{AMSTUTZ2006573}
\bibinfo{author}{S.~Amstutz}, \bibinfo{author}{H.~Andrä},
\newblock \bibinfo{title}{A new algorithm for topology optimization using a
  level-set method},
\newblock \bibinfo{journal}{Journal of Computational Physics}
  \bibinfo{volume}{216} (\bibinfo{year}{2006}) \bibinfo{pages}{573--588}.
  \URLprefix
  \url{https://www.sciencedirect.com/science/article/pii/S0021999105005656}.
  \DOIprefix\doi{https://doi.org/10.1016/j.jcp.2005.12.015}.
\bibitem[{Allaire et~al.(2004)Allaire, Jouve, and Toader}]{ALLAIRE2004363}
\bibinfo{author}{G.~Allaire}, \bibinfo{author}{F.~Jouve},
  \bibinfo{author}{A.-M. Toader},
\newblock \bibinfo{title}{Structural optimization using sensitivity analysis
  and a level-set method},
\newblock \bibinfo{journal}{Journal of Computational Physics}
  \bibinfo{volume}{194} (\bibinfo{year}{2004}) \bibinfo{pages}{363--393}.
  \URLprefix
  \url{https://www.sciencedirect.com/science/article/pii/S002199910300487X}.
  \DOIprefix\doi{https://doi.org/10.1016/j.jcp.2003.09.032}.
\bibitem[{Allard and Shahbazian(2014)}]{allard2014unmanned}
\bibinfo{author}{Y.~Allard}, \bibinfo{author}{E.~Shahbazian},
  \bibinfo{title}{Unmanned underwater vehicle (UUV) information study},
  \bibinfo{type}{Technical Report}, OODA Technologies Inc Montreal, Quebec
  Canada, \bibinfo{year}{2014}.
\bibitem[{Fletcher(2000)}]{fletcher2000uuv}
\bibinfo{author}{B.~Fletcher},
\newblock \bibinfo{title}{Uuv master plan: a vision for navy uuv development},
\newblock in: \bibinfo{booktitle}{OCEANS 2000 MTS/IEEE Conference and
  Exhibition. Conference Proceedings (Cat. No. 00CH37158)},
  volume~\bibinfo{volume}{1}, \bibinfo{organization}{IEEE},
  \bibinfo{year}{2000}, pp. \bibinfo{pages}{65--71}.
\bibitem[{Button et~al.(2009)Button, Kamp, Curtin, and
  Dryden}]{button2009survey}
\bibinfo{author}{R.~W. Button}, \bibinfo{author}{J.~Kamp},
  \bibinfo{author}{T.~B. Curtin}, \bibinfo{author}{J.~Dryden},
  \bibinfo{title}{A survey of missions for unmanned undersea vehicles},
  \bibinfo{type}{Technical Report}, {RAND} National Defense Research Institute,
  \bibinfo{year}{2009}.
\bibitem[{Liu et~al.(2021)Liu, Yu, Zhang, Liu, Feng, and Zhang}]{liu2021fine}
\bibinfo{author}{Y.~Liu}, \bibinfo{author}{Z.~Yu}, \bibinfo{author}{L.~Zhang},
  \bibinfo{author}{T.~Liu}, \bibinfo{author}{D.~Feng},
  \bibinfo{author}{J.~Zhang},
\newblock \bibinfo{title}{A fine drag coefficient model for hull shape of
  underwater vehicles},
\newblock \bibinfo{journal}{Ocean Engineering} \bibinfo{volume}{236}
  (\bibinfo{year}{2021}) \bibinfo{pages}{109361}.
\bibitem[{Myring(1976)}]{myring1976theoretical}
\bibinfo{author}{D.~Myring},
\newblock \bibinfo{title}{A theoretical study of body drag in subcritical
  axisymmetric flow},
\newblock \bibinfo{journal}{Aeronautical quarterly} \bibinfo{volume}{27}
  (\bibinfo{year}{1976}) \bibinfo{pages}{186--194}.
\bibitem[{French(2010)}]{french2010analysis}
\bibinfo{author}{D.~W. French}, \bibinfo{title}{Analysis of unmanned undersea
  vehicle (UUV) architectures and an assessment of UUV integration into
  undersea applications}, \bibinfo{type}{Technical Report}, NAVAL POSTGRADUATE
  SCHOOL MONTEREY CA, \bibinfo{year}{2010}.
\bibitem[{Crowell(2013)}]{crowell2013design}
\bibinfo{author}{J.~Crowell},
\newblock \bibinfo{title}{Design challenges of a next generation small auv},
\newblock in: \bibinfo{booktitle}{2013 OCEANS-San Diego},
  \bibinfo{organization}{IEEE}, \bibinfo{year}{2013}, pp.
  \bibinfo{pages}{1--5}.
\bibitem[{Sagaut et~al.(2013)Sagaut, Terracol, and Deck}]{sagaut2013multiscale}
\bibinfo{author}{P.~Sagaut}, \bibinfo{author}{M.~Terracol},
  \bibinfo{author}{S.~Deck}, \bibinfo{title}{Multiscale and multiresolution
  approaches in turbulence-LES, DES and Hybrid RANS/LES Methods: Applications
  and Guidelines}, \bibinfo{publisher}{World Scientific}, \bibinfo{year}{2013}.
\bibitem[{Menter(1992)}]{menter1992improved}
\bibinfo{author}{F.~R. Menter}, \bibinfo{title}{Improved two-equation k-omega
  turbulence models for aerodynamic flows}, \bibinfo{type}{Technical Report}
  \bibinfo{number}{103975}, NASA, \bibinfo{year}{1992}.
\bibitem[{Jasak et~al.(2007)Jasak, Jemcov, Tukovic et~al.}]{jasak2007openfoam}
\bibinfo{author}{H.~Jasak}, \bibinfo{author}{A.~Jemcov},
  \bibinfo{author}{Z.~Tukovic}, et~al.,
\newblock \bibinfo{title}{Openfoam: A c++ library for complex physics
  simulations},
\newblock in: \bibinfo{booktitle}{International workshop on coupled methods in
  numerical dynamics}, volume \bibinfo{volume}{1000},
  \bibinfo{organization}{IUC Dubrovnik Croatia}, \bibinfo{year}{2007}, pp.
  \bibinfo{pages}{1--20}.
\bibitem[{Gertler(1950)}]{gertler1950resistance}
\bibinfo{author}{M.~Gertler}, \bibinfo{title}{Resistance experiments on a
  systematic series of streamlined bodies of revolution: for application to the
  design of high-speed submarines}, \bibinfo{publisher}{Navy Department, David
  W. Taylor Model Basin}, \bibinfo{year}{1950}.
\bibitem[{Carmichael(1966)}]{carmichael1966underwater}
\bibinfo{author}{B.~H. Carmichael},
\newblock \bibinfo{title}{Underwater vehicle drag reduction through choice of
  shape},
\newblock in: \bibinfo{booktitle}{AIAA second propulsion joint specialist
  conference}, \bibinfo{year}{1966}.
\bibitem[{Neira et~al.(2021)Neira, Sequeiros, Huamani, Machaca, Fonseca, and
  Nina}]{neira2021review}
\bibinfo{author}{J.~Neira}, \bibinfo{author}{C.~Sequeiros},
  \bibinfo{author}{R.~Huamani}, \bibinfo{author}{E.~Machaca},
  \bibinfo{author}{P.~Fonseca}, \bibinfo{author}{W.~Nina},
\newblock \bibinfo{title}{Review on unmanned underwater robotics, structure
  designs, materials, sensors, actuators, and navigation control},
\newblock \bibinfo{journal}{Journal of Robotics} \bibinfo{volume}{2021}
  (\bibinfo{year}{2021}).
\bibitem[{Manley(2016)}]{manley2016unmanned}
\bibinfo{author}{J.~E. Manley},
\newblock \bibinfo{title}{Unmanned maritime vehicles, 20 years of commercial
  and technical evolution},
\newblock in: \bibinfo{booktitle}{OCEANS 2016 MTS/IEEE Monterey},
  \bibinfo{organization}{IEEE}, \bibinfo{year}{2016}, pp.
  \bibinfo{pages}{1--6}.
\bibitem[{Parsons et~al.(1974)Parsons, Goodson, and
  Goldschmied}]{parsons1974shaping}
\bibinfo{author}{J.~S. Parsons}, \bibinfo{author}{R.~E. Goodson},
  \bibinfo{author}{F.~R. Goldschmied},
\newblock \bibinfo{title}{Shaping of axisymmetric bodies for minimum drag in
  incompressible flow},
\newblock \bibinfo{journal}{Journal of Hydronautics} \bibinfo{volume}{8}
  (\bibinfo{year}{1974}) \bibinfo{pages}{100--107}.
\bibitem[{Hertel(1966)}]{hertel1966full}
\bibinfo{author}{H.~Hertel},
\newblock \bibinfo{title}{Full integration of vtol power plants in the aircraft
  fuselage(full integration of vtol lifting fans into aircraft fuselage)},
\newblock \bibinfo{journal}{1966.}  (\bibinfo{year}{1966})
  \bibinfo{pages}{65--96}.
\bibitem[{Hess(1976)}]{hess1976problem}
\bibinfo{author}{J.~L. Hess},
\newblock \bibinfo{title}{On the problem of shaping an axisymmetric body to
  obtain low drag at large reynolds numbers},
\newblock \bibinfo{journal}{Journal of Ship Research} \bibinfo{volume}{20}
  (\bibinfo{year}{1976}) \bibinfo{pages}{51--60}.
\bibitem[{Zedan and Dalton(1979)}]{zedan1979viscious}
\bibinfo{author}{M.~F. Zedan}, \bibinfo{author}{C.~Dalton},
\newblock \bibinfo{title}{Viscious drag computation for axisymmetric bodies at
  high reynolds numbers},
\newblock \bibinfo{journal}{Journal of Hydronautics} \bibinfo{volume}{13}
  (\bibinfo{year}{1979}) \bibinfo{pages}{52--60}.
\bibitem[{Lutz and Wagner(1998)}]{lutz1998drag}
\bibinfo{author}{T.~Lutz}, \bibinfo{author}{S.~Wagner},
\newblock \bibinfo{title}{Drag reduction and shape optimization of airship
  bodies},
\newblock \bibinfo{journal}{Journal of Aircraft} \bibinfo{volume}{35}
  (\bibinfo{year}{1998}) \bibinfo{pages}{345--351}.
\bibitem[{Alvarez et~al.(2009)Alvarez, Bertram, and Gualdesi}]{alvarez2009hull}
\bibinfo{author}{A.~Alvarez}, \bibinfo{author}{V.~Bertram},
  \bibinfo{author}{L.~Gualdesi},
\newblock \bibinfo{title}{Hull hydrodynamic optimization of autonomous
  underwater vehicles operating at snorkeling depth},
\newblock \bibinfo{journal}{Ocean Engineering} \bibinfo{volume}{36}
  (\bibinfo{year}{2009}) \bibinfo{pages}{105--112}.
\bibitem[{Stevenson et~al.(2007)Stevenson, Furlong, and
  Dormer}]{stevenson2007auv}
\bibinfo{author}{P.~Stevenson}, \bibinfo{author}{M.~Furlong},
  \bibinfo{author}{D.~Dormer},
\newblock \bibinfo{title}{{AUV} shapes-combining the practical and hydrodynamic
  considerations},
\newblock in: \bibinfo{booktitle}{Oceans 2007-Europe},
  \bibinfo{organization}{IEEE}, \bibinfo{year}{2007}, pp.
  \bibinfo{pages}{1--6}.
\bibitem[{Zifan et~al.(2014)Zifan, Qiang, and Songlin}]{zifan2014analysis}
\bibinfo{author}{W.~Zifan}, \bibinfo{author}{Y.~Qiang},
  \bibinfo{author}{Y.~Songlin},
\newblock \bibinfo{title}{Analysis of the resistance performance for different
  types of auvs based on cfd},
\newblock \bibinfo{journal}{Chinese Journal of Ship Research}
  \bibinfo{volume}{9} (\bibinfo{year}{2014}) \bibinfo{pages}{28--37}.
\bibitem[{Yamamoto(2015)}]{yamamoto2015research}
\bibinfo{author}{I.~Yamamoto},
\newblock \bibinfo{title}{Research on next autonomous underwater vehicle for
  longer distance cruising},
\newblock \bibinfo{journal}{IFAC-PapersOnLine} \bibinfo{volume}{48}
  (\bibinfo{year}{2015}) \bibinfo{pages}{173--176}.
\bibitem[{Alam et~al.(2015)Alam, Ray, and Anavatti}]{alam2015design}
\bibinfo{author}{K.~Alam}, \bibinfo{author}{T.~Ray}, \bibinfo{author}{S.~G.
  Anavatti},
\newblock \bibinfo{title}{Design optimization of an unmanned underwater vehicle
  using low-and high-fidelity models},
\newblock \bibinfo{journal}{IEEE Transactions on Systems, Man, and Cybernetics:
  Systems} \bibinfo{volume}{47} (\bibinfo{year}{2015})
  \bibinfo{pages}{2794--2808}.
\bibitem[{Jameson(2003)}]{jameson2003aerodynamic}
\bibinfo{author}{A.~Jameson},
\newblock \bibinfo{title}{Aerodynamic shape optimization using the adjoint
  method},
\newblock \bibinfo{journal}{Lectures at the Von Karman Institute, Brussels}
  (\bibinfo{year}{2003}).
\bibitem[{Song et~al.(2010)Song, Zhu, and Liu}]{song2010research}
\bibinfo{author}{B.~Song}, \bibinfo{author}{Q.~Zhu}, \bibinfo{author}{Z.~Liu},
\newblock \bibinfo{title}{Research on multi-objective optimization design of
  the uuv shape based on numerical simulation},
\newblock in: \bibinfo{booktitle}{International Conference in Swarm
  Intelligence}, \bibinfo{organization}{Springer}, \bibinfo{year}{2010}, pp.
  \bibinfo{pages}{628--635}.
\bibitem[{Schweyher et~al.(1996)Schweyher, Lutz, and
  Wagner}]{schweyher1996optimization}
\bibinfo{author}{H.~Schweyher}, \bibinfo{author}{T.~Lutz},
  \bibinfo{author}{S.~Wagner},
\newblock \bibinfo{title}{An optimization tool for axisymmetric bodies of
  minimum drag},
\newblock in: \bibinfo{booktitle}{2nd international airship conference,
  Stuttgart/Friedrichshafen}, \bibinfo{year}{1996}, pp. \bibinfo{pages}{3--4}.
\bibitem[{Vardhan and Sztipanovits(2023)}]{vardhan2023search}
\bibinfo{author}{H.~Vardhan}, \bibinfo{author}{J.~Sztipanovits},
\newblock \bibinfo{title}{Search for universal minimum drag resistance
  underwater vehicle hull using cfd},
\newblock \bibinfo{journal}{arXiv preprint arXiv:2302.09441}
  (\bibinfo{year}{2023}).
\bibitem[{Eismann et~al.(2017)Eismann, Bartzsch, and Ermon}]{eismann2017shape}
\bibinfo{author}{S.~Eismann}, \bibinfo{author}{S.~Bartzsch},
  \bibinfo{author}{S.~Ermon},
\newblock \bibinfo{title}{Shape optimization in laminar flow with a
  label-guided variational autoencoder},
\newblock \bibinfo{journal}{arXiv preprint arXiv:1712.03599}
  (\bibinfo{year}{2017}).
\bibitem[{Vardhan et~al.(2023)Vardhan, Volgyesi, and
  Sztipanovits}]{vardhan2023constrained}
\bibinfo{author}{H.~Vardhan}, \bibinfo{author}{P.~Volgyesi},
  \bibinfo{author}{J.~Sztipanovits},
\newblock \bibinfo{title}{Constrained bayesian optimization for automatic
  underwater vehicle hull design},
\newblock \bibinfo{journal}{arXiv preprint arXiv:2302.14732}
  (\bibinfo{year}{2023}).
\bibitem[{Vardhan et~al.(2022)Vardhan, Timalsina, Volgyesi, and
  Sztipanovits}]{vardhan2022data}
\bibinfo{author}{H.~Vardhan}, \bibinfo{author}{U.~Timalsina},
  \bibinfo{author}{P.~Volgyesi}, \bibinfo{author}{J.~Sztipanovits},
\newblock \bibinfo{title}{Data efficient surrogate modeling for engineering
  design: Ensemble-free batch mode deep active learning for regression},
\newblock \bibinfo{journal}{arXiv preprint arXiv:2211.10360}
  (\bibinfo{year}{2022}).
\bibitem[{Rasmussen(2003)}]{rasmussen2003gaussian}
\bibinfo{author}{C.~E. Rasmussen},
\newblock \bibinfo{title}{Gaussian processes in machine learning},
\newblock in: \bibinfo{booktitle}{Summer school on machine learning},
  \bibinfo{organization}{Springer}, \bibinfo{year}{2003}, pp.
  \bibinfo{pages}{63--71}.
\bibitem[{Forrester et~al.(2008)Forrester, Sobester, and
  Keane}]{forrester2008engineering}
\bibinfo{author}{A.~Forrester}, \bibinfo{author}{A.~Sobester},
  \bibinfo{author}{A.~Keane}, \bibinfo{title}{Engineering design via surrogate
  modelling: a practical guide}, \bibinfo{publisher}{John Wiley \& Sons},
  \bibinfo{year}{2008}.
\bibitem[{Marsden et~al.(2004)Marsden, Wang, Dennis, and
  Moin}]{marsden2004optimal}
\bibinfo{author}{A.~L. Marsden}, \bibinfo{author}{M.~Wang},
  \bibinfo{author}{J.~E. Dennis}, \bibinfo{author}{P.~Moin},
\newblock \bibinfo{title}{Optimal aeroacoustic shape design using the surrogate
  management framework},
\newblock \bibinfo{journal}{Optimization and Engineering} \bibinfo{volume}{5}
  (\bibinfo{year}{2004}) \bibinfo{pages}{235--262}.
\bibitem[{Abouhussein and Peet(2023)}]{abouhussein2023computational}
\bibinfo{author}{A.~Abouhussein}, \bibinfo{author}{Y.~T. Peet},
\newblock \bibinfo{title}{Computational framework for efficient high-fidelity
  optimization of bio-inspired propulsion and its application to accelerating
  swimmers},
\newblock \bibinfo{journal}{Journal of Computational Physics}
  \bibinfo{volume}{482} (\bibinfo{year}{2023}) \bibinfo{pages}{112038}.
\bibitem[{Booker et~al.(1999)Booker, Dennis, Frank, Serafini, Torczon, and
  Trosset}]{booker1999rigorous}
\bibinfo{author}{A.~J. Booker}, \bibinfo{author}{J.~E. Dennis},
  \bibinfo{author}{P.~D. Frank}, \bibinfo{author}{D.~B. Serafini},
  \bibinfo{author}{V.~Torczon}, \bibinfo{author}{M.~W. Trosset},
\newblock \bibinfo{title}{A rigorous framework for optimization of expensive
  functions by surrogates},
\newblock \bibinfo{journal}{Structural optimization} \bibinfo{volume}{17}
  (\bibinfo{year}{1999}) \bibinfo{pages}{1--13}.
\bibitem[{Morita et~al.(2022)Morita, Rezaeiravesh, Tabatabaei, Vinuesa,
  Fukagata, and Schlatter}]{morita2022applying}
\bibinfo{author}{Y.~Morita}, \bibinfo{author}{S.~Rezaeiravesh},
  \bibinfo{author}{N.~Tabatabaei}, \bibinfo{author}{R.~Vinuesa},
  \bibinfo{author}{K.~Fukagata}, \bibinfo{author}{P.~Schlatter},
\newblock \bibinfo{title}{Applying bayesian optimization with gaussian process
  regression to computational fluid dynamics problems},
\newblock \bibinfo{journal}{Journal of Computational Physics}
  \bibinfo{volume}{449} (\bibinfo{year}{2022}) \bibinfo{pages}{110788}.
\bibitem[{Bhatnagar et~al.(2019)Bhatnagar, Afshar, Pan, Duraisamy, and
  Kaushik}]{bhatnagar2019prediction}
\bibinfo{author}{S.~Bhatnagar}, \bibinfo{author}{Y.~Afshar},
  \bibinfo{author}{S.~Pan}, \bibinfo{author}{K.~Duraisamy},
  \bibinfo{author}{S.~Kaushik},
\newblock \bibinfo{title}{Prediction of aerodynamic flow fields using
  convolutional neural networks},
\newblock \bibinfo{journal}{Computational Mechanics} \bibinfo{volume}{64}
  (\bibinfo{year}{2019}) \bibinfo{pages}{525--545}.
\bibitem[{Chen et~al.(2019)Chen, Viquerat, and Hachem}]{chen2019u}
\bibinfo{author}{J.~Chen}, \bibinfo{author}{J.~Viquerat},
  \bibinfo{author}{E.~Hachem},
\newblock \bibinfo{title}{U-net architectures for fast prediction of
  incompressible laminar flows},
\newblock \bibinfo{journal}{arXiv preprint arXiv:1910.13532}
  (\bibinfo{year}{2019}).
\bibitem[{Ronneberger et~al.(2015)Ronneberger, Fischer, and
  Brox}]{ronneberger2015u}
\bibinfo{author}{O.~Ronneberger}, \bibinfo{author}{P.~Fischer},
  \bibinfo{author}{T.~Brox},
\newblock \bibinfo{title}{U-net: Convolutional networks for biomedical image
  segmentation},
\newblock in: \bibinfo{booktitle}{International Conference on Medical image
  computing and computer-assisted intervention},
  \bibinfo{organization}{Springer}, \bibinfo{year}{2015}, pp.
  \bibinfo{pages}{234--241}.
\bibitem[{Vardhan et~al.(2021)Vardhan, Volgyesi, and
  Sztipanovits}]{vardhan2021machine}
\bibinfo{author}{H.~Vardhan}, \bibinfo{author}{P.~Volgyesi},
  \bibinfo{author}{J.~Sztipanovits},
\newblock \bibinfo{title}{Machine learning assisted propeller design},
\newblock in: \bibinfo{booktitle}{Proceedings of the ACM/IEEE 12th
  International Conference on Cyber-Physical Systems}, \bibinfo{year}{2021},
  pp. \bibinfo{pages}{227--228}.
\bibitem[{Vardhan and Sztipanovits(2022)}]{vardhan2022deep}
\bibinfo{author}{H.~Vardhan}, \bibinfo{author}{J.~Sztipanovits},
\newblock \bibinfo{title}{Deep learning-based fea surrogate for sub-sea
  pressure vessel},
\newblock \bibinfo{journal}{arXiv preprint arXiv:2206.03322}
  (\bibinfo{year}{2022}).
\bibitem[{Jones et~al.(2016)Jones, Chapuis, Liefvendahl, Norrison, and
  Widjaja}]{jones2016rans}
\bibinfo{author}{D.~A. Jones}, \bibinfo{author}{M.~Chapuis},
  \bibinfo{author}{M.~Liefvendahl}, \bibinfo{author}{D.~Norrison},
  \bibinfo{author}{R.~Widjaja}, \bibinfo{title}{{RANS} simulations using
  {OpenFOAM} software}, \bibinfo{type}{Technical Report}, Defence Science and
  Technology Group Fishermans Bend Victoria Australia, \bibinfo{year}{2016}.
\bibitem[{Wilcox et~al.(1998)}]{wilcox1998turbulence}
\bibinfo{author}{D.~C. Wilcox}, et~al., \bibinfo{title}{Turbulence modeling for
  CFD}, volume~\bibinfo{volume}{2}, \bibinfo{publisher}{DCW industries La
  Canada, CA}, \bibinfo{year}{1998}.
\bibitem[{Launder and Spalding(1983)}]{launder1983numerical}
\bibinfo{author}{B.~E. Launder}, \bibinfo{author}{D.~B. Spalding},
\newblock \bibinfo{title}{The numerical computation of turbulent flows},
\newblock in: \bibinfo{booktitle}{Numerical prediction of flow, heat transfer,
  turbulence and combustion}, \bibinfo{publisher}{Elsevier},
  \bibinfo{year}{1983}, pp. \bibinfo{pages}{96--116}.
\bibitem[{Menter(1992)}]{menter1992influence}
\bibinfo{author}{F.~R. Menter},
\newblock \bibinfo{title}{Influence of freestream values on k-omega turbulence
  model predictions},
\newblock \bibinfo{journal}{AIAA journal} \bibinfo{volume}{30}
  (\bibinfo{year}{1992}) \bibinfo{pages}{1657--1659}.
\bibitem[{Tide and Babu(2008)}]{tide2008comparison}
\bibinfo{author}{P.~Tide}, \bibinfo{author}{V.~Babu},
\newblock \bibinfo{title}{A comparison of predictions by sst and wilcox kw
  models for a mach 0.9 jet},
\newblock in: \bibinfo{booktitle}{46th AIAA Aerospace Sciences Meeting and
  Exhibit}, \bibinfo{year}{2008}, p.~\bibinfo{pages}{24}.
\bibitem[{Nikiforow et~al.(2016)Nikiforow, Koski, Karim{\"a}ki, Ihonen, and
  Alopaeus}]{nikiforow2016designing}
\bibinfo{author}{K.~Nikiforow}, \bibinfo{author}{P.~Koski},
  \bibinfo{author}{H.~Karim{\"a}ki}, \bibinfo{author}{J.~Ihonen},
  \bibinfo{author}{V.~Alopaeus},
\newblock \bibinfo{title}{Designing a hydrogen gas ejector for 5 kw stationary
  pemfc system--cfd-modeling and experimental validation},
\newblock \bibinfo{journal}{International Journal of Hydrogen Energy}
  \bibinfo{volume}{41} (\bibinfo{year}{2016}) \bibinfo{pages}{14952--14970}.
\bibitem[{Riegel et~al.(2016)Riegel, Mayer, and van Havre}]{riegel2016freecad}
\bibinfo{author}{J.~Riegel}, \bibinfo{author}{W.~Mayer},
  \bibinfo{author}{Y.~van Havre}, \bibinfo{title}{Freecad},
  \bibinfo{year}{2016}.
\bibitem[{authors(2016)}]{gpyopt2016}
\bibinfo{author}{T.~G. authors}, \bibinfo{title}{{GPyOpt}: A bayesian
  optimization framework in python},
  \bibinfo{howpublished}{\url{http://github.com/SheffieldML/GPyOpt}},
  \bibinfo{year}{2016}.
\bibitem[{Blank and Deb(2020)}]{blank2020pymoo}
\bibinfo{author}{J.~Blank}, \bibinfo{author}{K.~Deb},
\newblock \bibinfo{title}{Pymoo: Multi-objective optimization in python},
\newblock \bibinfo{journal}{IEEE Access} \bibinfo{volume}{8}
  (\bibinfo{year}{2020}) \bibinfo{pages}{89497--89509}.
\bibitem[{Hoffmann and Kim(2001)}]{hoffmann2001towards}
\bibinfo{author}{C.~M. Hoffmann}, \bibinfo{author}{K.-J. Kim},
\newblock \bibinfo{title}{Towards valid parametric cad models},
\newblock \bibinfo{journal}{Computer-Aided Design} \bibinfo{volume}{33}
  (\bibinfo{year}{2001}) \bibinfo{pages}{81--90}.
\bibitem[{Patankar and Spalding(1983)}]{patankar1983calculation}
\bibinfo{author}{S.~V. Patankar}, \bibinfo{author}{D.~B. Spalding},
\newblock \bibinfo{title}{A calculation procedure for heat, mass and momentum
  transfer in three-dimensional parabolic flows},
\newblock in: \bibinfo{booktitle}{Numerical prediction of flow, heat transfer,
  turbulence and combustion}, \bibinfo{publisher}{Elsevier},
  \bibinfo{year}{1983}, pp. \bibinfo{pages}{54--73}.
\bibitem[{Ayachit(2015)}]{ayachit2015paraview}
\bibinfo{author}{U.~Ayachit}, \bibinfo{title}{The {ParaView} guide: a parallel
  visualization application}, \bibinfo{publisher}{Kitware, Inc.},
  \bibinfo{year}{2015}.
\bibitem[{Gao et~al.(2016)Gao, Wang, Pang, and Cao}]{gao2016hull}
\bibinfo{author}{T.~Gao}, \bibinfo{author}{Y.~Wang}, \bibinfo{author}{Y.~Pang},
  \bibinfo{author}{J.~Cao},
\newblock \bibinfo{title}{Hull shape optimization for autonomous underwater
  vehicles using cfd},
\newblock \bibinfo{journal}{Engineering applications of computational fluid
  mechanics} \bibinfo{volume}{10} (\bibinfo{year}{2016})
  \bibinfo{pages}{599--607}.
\bibitem[{Morris and Mitchell(1995)}]{morris1995exploratory}
\bibinfo{author}{M.~D. Morris}, \bibinfo{author}{T.~J. Mitchell},
\newblock \bibinfo{title}{Exploratory designs for computational experiments},
\newblock \bibinfo{journal}{Journal of statistical planning and inference}
  \bibinfo{volume}{43} (\bibinfo{year}{1995}) \bibinfo{pages}{381--402}.
\bibitem[{Holland(1992)}]{holland1992adaptation}
\bibinfo{author}{J.~H. Holland}, \bibinfo{title}{Adaptation in natural and
  artificial systems: an introductory analysis with applications to biology,
  control, and artificial intelligence}, \bibinfo{publisher}{MIT press},
  \bibinfo{year}{1992}.
\bibitem[{Vose(1999)}]{vose1999simple}
\bibinfo{author}{M.~D. Vose}, \bibinfo{title}{The simple genetic algorithm:
  foundations and theory}, \bibinfo{publisher}{MIT press},
  \bibinfo{year}{1999}.
\bibitem[{Nelder and Mead(1965)}]{nelder1965simplex}
\bibinfo{author}{J.~A. Nelder}, \bibinfo{author}{R.~Mead},
\newblock \bibinfo{title}{A simplex method for function minimization},
\newblock \bibinfo{journal}{The computer journal} \bibinfo{volume}{7}
  (\bibinfo{year}{1965}) \bibinfo{pages}{308--313}.
\bibitem[{Clark(1961)}]{clark1961greatest}
\bibinfo{author}{C.~E. Clark},
\newblock \bibinfo{title}{The greatest of a finite set of random variables},
\newblock \bibinfo{journal}{Operations Research} \bibinfo{volume}{9}
  (\bibinfo{year}{1961}) \bibinfo{pages}{145--162}.
\bibitem[{Kushner(1964)}]{kushner1964new}
\bibinfo{author}{H.~J. Kushner},
\newblock \bibinfo{title}{A new method of locating the maximum point of an
  arbitrary multipeak curve in the presence of noise}  (\bibinfo{year}{1964}).
\bibitem[{Zhilinskas(1975)}]{zhilinskas1975single}
\bibinfo{author}{A.~Zhilinskas},
\newblock \bibinfo{title}{Single-step bayesian search method for an extremum of
  functions of a single variable},
\newblock \bibinfo{journal}{Cybernetics} \bibinfo{volume}{11}
  (\bibinfo{year}{1975}) \bibinfo{pages}{160--166}.
\bibitem[{Mo{\v{c}}kus(1975)}]{movckus1975bayesian}
\bibinfo{author}{J.~Mo{\v{c}}kus},
\newblock \bibinfo{title}{On bayesian methods for seeking the extremum},
\newblock in: \bibinfo{booktitle}{Optimization techniques IFIP technical
  conference}, \bibinfo{organization}{Springer}, \bibinfo{year}{1975}, pp.
  \bibinfo{pages}{400--404}.
\bibitem[{Jones et~al.(1998)Jones, Schonlau, and Welch}]{jones1998efficient}
\bibinfo{author}{D.~R. Jones}, \bibinfo{author}{M.~Schonlau},
  \bibinfo{author}{W.~J. Welch},
\newblock \bibinfo{title}{Efficient global optimization of expensive black-box
  functions},
\newblock \bibinfo{journal}{Journal of Global optimization}
  \bibinfo{volume}{13} (\bibinfo{year}{1998}) \bibinfo{pages}{455--492}.
\bibitem[{Srinivas et~al.(2012)Srinivas, Krause, Kakade, and
  Seeger}]{srinivas2012information}
\bibinfo{author}{N.~Srinivas}, \bibinfo{author}{A.~Krause},
  \bibinfo{author}{S.~M. Kakade}, \bibinfo{author}{M.~W. Seeger},
\newblock \bibinfo{title}{Information-theoretic regret bounds for gaussian
  process optimization in the bandit setting},
\newblock \bibinfo{journal}{IEEE transactions on information theory}
  \bibinfo{volume}{58} (\bibinfo{year}{2012}) \bibinfo{pages}{3250--3265}.
\bibitem[{Winey(2020)}]{winey2020modifiable}
\bibinfo{author}{N.~E. Winey}, \bibinfo{title}{Modifiable stability and
  maneuverability of high speed unmanned underwater vehicles (UUVs) through
  bioinspired control fins}, Ph.D. thesis, Massachusetts Institute of
  Technology, \bibinfo{year}{2020}.
\bibitem[{Xu et~al.(2015)Xu, Wang, Chen, and Li}]{xu2015empirical}
\bibinfo{author}{B.~Xu}, \bibinfo{author}{N.~Wang}, \bibinfo{author}{T.~Chen},
  \bibinfo{author}{M.~Li},
\newblock \bibinfo{title}{Empirical evaluation of rectified activations in
  convolutional network},
\newblock \bibinfo{journal}{arXiv preprint arXiv:1505.00853}
  (\bibinfo{year}{2015}).
\bibitem[{He et~al.(2016)He, Zhang, Ren, and Sun}]{he2016deep}
\bibinfo{author}{K.~He}, \bibinfo{author}{X.~Zhang}, \bibinfo{author}{S.~Ren},
  \bibinfo{author}{J.~Sun},
\newblock \bibinfo{title}{Deep residual learning for image recognition},
\newblock in: \bibinfo{booktitle}{Proceedings of the IEEE conference on
  computer vision and pattern recognition}, \bibinfo{year}{2016}, pp.
  \bibinfo{pages}{770--778}.
\bibitem[{Kingma and Ba(2014)}]{kingma2014adam}
\bibinfo{author}{D.~P. Kingma}, \bibinfo{author}{J.~Ba},
\newblock \bibinfo{title}{Adam: A method for stochastic optimization},
\newblock \bibinfo{journal}{arXiv preprint arXiv:1412.6980}
  (\bibinfo{year}{2014}).
\bibitem[{Glorot and Bengio(2010)}]{glorot2010understanding}
\bibinfo{author}{X.~Glorot}, \bibinfo{author}{Y.~Bengio},
\newblock \bibinfo{title}{Understanding the difficulty of training deep
  feedforward neural networks},
\newblock in: \bibinfo{booktitle}{Proceedings of the thirteenth international
  conference on artificial intelligence and statistics},
  \bibinfo{organization}{JMLR Workshop and Conference Proceedings},
  \bibinfo{year}{2010}, pp. \bibinfo{pages}{249--256}.
\bibitem[{Bradford et~al.(2018)Bradford, Schweidtmann, and
  Lapkin}]{bradford2018efficient}
\bibinfo{author}{E.~Bradford}, \bibinfo{author}{A.~M. Schweidtmann},
  \bibinfo{author}{A.~Lapkin},
\newblock \bibinfo{title}{Efficient multiobjective optimization employing
  gaussian processes, spectral sampling and a genetic algorithm},
\newblock \bibinfo{journal}{Journal of global optimization}
  \bibinfo{volume}{71} (\bibinfo{year}{2018}) \bibinfo{pages}{407--438}.
\bibitem[{Choi et~al.(2006)Choi, Jeon, and Choi}]{choi2006mechanism}
\bibinfo{author}{J.~Choi}, \bibinfo{author}{W.-P. Jeon},
  \bibinfo{author}{H.~Choi},
\newblock \bibinfo{title}{Mechanism of drag reduction by dimples on a sphere},
\newblock \bibinfo{journal}{Physics of Fluids} \bibinfo{volume}{18}
  (\bibinfo{year}{2006}) \bibinfo{pages}{041702}.

\end{thebibliography}

\appendix
\section{Sampling and optimization methods}
\label{sec:optimization-methods}

In this section, we provide a background on the different sampling and optimization techniques used in this work.  Optimization frameworks are generally classified as gradient-based and gradient-free methods.  Since the gradient information in finite volume-based CFD simulation is very costly to compute and unreliable (even using adjoint solvers to compute gradients, and computing the adjoint sensitivities could be expensive), we chose gradient-free optimization frameworks for this study.

\subsection{Bayesian Optimization}
Bayesian optimization starts with a probabilistic belief (Gaussian prior) about the function $f$ and conditions this belief on new observations. During the entire optimization cycle, BO maintains this probabilistic model about $f$. 
 $$p(f)= \mathcal{GP}(f; \mu,K)$$
 Given observations $D=(X,f),$ we can condition our distribution on D as usual: 
 $$p(f|D)= \mathcal{GP}(f; \mu_{f|D},K_{f|D})$$
 
By maintaining a probabilistic belief about the function $f$, an acquisition function $a(x)$)can be designed to determine where to evaluate the function next.
The acquisition function is typically an inexpensive function that can be evaluated at different points domain using $p(f|D)$ to measure how desirable evaluating $f$ at $x$ is expected to be for optimization purposes. We then optimize the acquisition function to select the most desirable location for the next observation. Alternatively, in the framework of Bayesian decision theory,  $a()$ can be interpreted as evaluating an expected regret associated with evaluating $f$ at a point $x$. The aim of optimization of the acquisition function is to select the point that can bring maximum improvement in the optimization objective.

\subsubsection{Expected Improvement} If $p(f|D)$ is the current belief about the function $f$ and $f'$ is the minimal value of $f$ observed so far, the EI evaluates the function $f$ at the point that, in expectation, improves the most upon $f'$. The utility function in such a case would be: 
$$u(x)=\max(0,f'-f(x))$$
Once evaluated, the reward would be $f'-f(x)$ if the value of the function at a newly evaluated sample $x$ (i.e., $f(x)$) is less than the current minimum $f'$, else it would be $0$. Then, the expected improvement using this utility function as a function of $x$ would be: 
\begin{align*}
    a_{EI}(x) = \mathbb{E}[u(x)|x,D]&= \int^{f'}_{-\infty} (f'-f) \mathcal{N}(f;\mu(x),K(x,x))df \\
    &= (f'-\mu(x)))\psi(f';\mu(x),K(x,x))+K(x,x)\mathcal{N}(f';\mu(x),K(x,x)) .
\end{align*}

This expected improvement of utility  function attempts to reduce the mean
function $\mu(x)$ (explicitly encoding exploitation by evaluating at points
with low mean) and also reduce the variance $K(x,x)$ (exploration by evaluating at points with high uncertainty). EI covers both exploration and exploitation in a Bayesian decision-theoretic way. 

\subsubsection{Lower Confidence bound} The acquisition function in LCB takes the form of
$$  a_{LCB}(x;\beta)= \mu(x)-\beta \sigma(x) .$$
Here, $\beta \geq 0$ is a hyperparameter that trades off exploration and exploitation in the optimization process.
Marginal standard deviation $\sigma(x)$ of the function $f(x)$ is defined as $\sqrt{K(x,x)}$.
An alternative view of the above acquisition function has been
proposed by Srinivas et al.\ \cite{srinivas2012information}, formulating the Bayesian optimization problem
as a multi-armed bandit and acquisition as an instantaneous regret function ($r(.)$). In such a case,
$$r(x)= f'-f(x) ,$$
and the goal of optimization in this framework is to find
$$\min  \sum_{t}^{T} r(x_t) = \max  \sum_{t}^{T} f(x_t) ,$$
where $T$ is the budget of evaluation for the optimization process. The hyperparameter $\beta$ in this work is changed to $\sqrt{(v,\tau_t)}$, with $v=1$ and $T_t = 2 \log(t^{d/2+2} \pi^2/ 3\delta)$ has shown no regret with high probability, i.e. $\lim_{T \rightarrow \infty} R_T/T =0$, where $R_T$ is cumulative regret ($R_T =\sum_{t}^{T} r(x_t)$).

\subsection{Genetic Algorithm}
GA involves the encoding of an optimization function as arrays of bits
or character strings to represent chromosomes, the manipulation operations of strings
by genetic operators, and the selection according to their fitness, with the aim to find a good (even optimal) solution to the problem concerned.
This is often done by the following procedure: (1) encoding the objectives or cost
functions; (2) defining a fitness function or selection criterion; (3) creating a population of individuals; (4) carrying out the evolution cycle or iterations by evaluating the
fitness of all the individuals in the population, creating a new population by performing
crossover and mutation, fitness-proportionate reproduction, etc., and replacing the old
population and iterating again using the new population; (5) decoding the results to
obtain the solution to the problem. These steps can be represented schematically as the
pseudo-code of genetic algorithms shown below.
One iteration of creating a new population is called a generation.
\begin{algorithm}[!ht]
\caption{GA pseudo-code}\label{alg:ga}
\begin{algorithmic}
\State Objective function: $f(x)$ 
\State Initialization: crossover probability ($p_c$) and mutation probability ($p_m$)
\State Encode the solutions ($x$) in chromosomes (strings)
\State Define fitness $F$ ($F \propto 1/f(x)$ ) and generate initial population and evaluate (randomly)
\While {$t < num \;generations$}
  \State Generate new solution by crossover
\State Mutate these solutions 
\State Accept the new solutions if their fitness increase
\State Select the current best for the next generation (elitism)
\State Update t = t + 1
\EndWhile
\end{algorithmic}
\end{algorithm}

\subsection{Nelder-Mead}
Nelder-Mead is a simplex-based direct search algorithm for a multidimensional unconstrained optimization problem. A simplex $S \in \mathbb{R}^n$ is the convex hull of $n+1$ vertices $S= \{v_i \}_{i=1,n+1}$. The algorithm is based on an iterative update on simplex size and shape by changing a vertex based on the associated function value
$f_i = f(v_i)$ for $i = 1,2,...,n+1$. The centroid of the simplex $c(j)$ is the center of the vertices where the vertex $v_j$ has been excluded:
$$c(j)= \frac{1}{n} \sum_{i=1,n+1,i \neq j} v_i .$$
Since it is a minimization problem, the vertices are sorted by decreasing function values $f(v_i)$ so that the best vertex has index 1 and the worst vertex has index $n + 1$. On a chosen hyperparameter (reflection factor $\rho >0$), then at each iteration, the algorithm tries to replace vertex $v_j$ by a new vertex $v_{\text{new}}(p,j)$ and on the line passing from vertex $v_j$ to centroid $c(j)$,
$$v_{\text{new}}(\rho,j)= (1+\rho)c(j)-\rho v(j)  .$$

The algorithm includes three other parameters of the algorithm: the coefficient of expansion ($\chi$), contraction ($\gamma$), and shrinkage ($\sigma$).  These are used when expansion or contraction steps are performed to change the shape of the simplex to adapt the local landscape (refer to \cite{nelder1965simplex} for greater details).

\subsection{Maximin Latin Hyper Cube}
In design of experiments, 
Latin Hypercube Sampling (LHC) has a long history and very robust performance. The main property of standard LHC is by projecting an $n$-point
design on to any factor, we will get $n$ different levels for that factor, which is a better representation of the design space. But due to the combinatorial number of permutations involved, it is difficult to get 
good Latin hypercube samples.
The idea of LHC is extended to find the best design by optimizing a criterion that describes a desirable property of design  instead of using a permutation of the $n$-level factor. 
For this purpose, a space-filling criterion is added to vanilla LHC by Morris and Mitchell \cite{morris1995exploratory} with the optimization goal of maximizing the minimum distance between the points.
Let $X$ be the design matrix with size $n \times k$, and let $s$ and $t$ be two design points.  Then, the distance measured between these two points would be $dist(s,t)= \{\sum_{j=1}^{k}|s_j-t_j|^p\}^{1/p}$.
Different $p$-norms may be chosen.
Morris and Mitchell \cite{morris1995exploratory}  applied this criterion to LHC to find optimal LHC. Since there are many designs that maximize the minimum inter-sample distance, an alternative definition of the maximin criterion is proposed. For a
given LHC, a distance list $(d_1; d_2;...; d_m)$ is created in ascending order of distance measurement (the maximum value of $m$ could be $\binom m 2$).  For the distance list, corresponding sample pairs that are separated by $d_i$ are also created, called $J_i$. Then a design $X$ is
called a maximin design if it sequentially maximizes $d_i$ and minimizes $J_i$ in the following
order: $d_1, J_1, d_2, J_2, ..., d_m, J_m$. 
This process is achieved by a simplified scalar-valued function which can be used to rank competing designs in such a way that the maximin design receives the highest ranking. The family of functions indexed by $p$ is given by
\begin{equation}
  \phi_p= \Big(\sum_{i=1}^{m} \frac{J_i}{d_i^p}\Big)^{1/p} ; \text{where} \; p \in  \mathbbm{Z}^+   .
\end{equation}
For large $p$, the design that minimizes $\phi_p$ will be a maximin design.

\subsection{Monte Carlo}
Monte Carlo sampling is a computational technique to construct a random process on a given seed for a range and dimension of input. The technique can be carried out by $N$-fold sampling from a random sequence of numbers with a prescribed probability distribution. The prescribed distribution used in this work was uniformly random (i.e., assuming each experiment in the design space has equal likelihood).  Samples are thereby selected from a uniform probability distribution function (pdf) between $0$ to $1$ and scaled in each dimension to the given parameter range.

\end{document}